\documentclass{article}
\usepackage{ijcai22}

\usepackage{times}
\usepackage{soul}
\usepackage{url}
\usepackage[hidelinks]{hyperref}
\usepackage[utf8]{inputenc}
\usepackage[small]{caption}
\usepackage{graphicx}
\usepackage{amsmath}
\usepackage{amsthm}
\usepackage{booktabs}
\usepackage{algorithm}
\usepackage{algorithmic}
\usepackage{subfigure, multirow, array, tabularx, amsfonts, bm, xcolor}
\usepackage{adjustbox}
\usepackage{arydshln}
\usepackage{pifont}% http://ctan.org/pkg/pifont
\usepackage{capt-of}
\usepackage{float,wrapfig}
\usepackage{comment}
\usepackage{array}
\newcolumntype{P}[1]{>{\centering\arraybackslash}p{#1}}

\newcommand{\cmark}{\ding{51}} % checkmark
\newcommand{\xmark}{\ding{55}} % cross
\newcommand{\rpm}{\raisebox{.2ex}{$\scriptstyle\pm$}}
\newcolumntype{P}[1]{>{\centering\arraybackslash}p{#1}}

\title{Few-Shot Adaptation of Pre-Trained Networks for Domain Shift}

\author{
Wenyu Zhang$^1$
\and
Li Shen$^1$\and
Wanyue Zhang$^2$\footnote{Work done while at Institute for Infocomm Research.}\And
Chuan-Sheng Foo$^{1,3}$
\affiliations
$^1$Institute for Infocomm Research, A*STAR\\
$^2$Max Planck Institute for Informatics\\
$^3$Centre for Frontier AI Research, A*STAR\\
\emails
\{zhang\_wenyu, lshen\}@i2r.a-star.edu.sg,
wzhang@mpi-inf.mpg.de,
foo\_chuan\_sheng@i2r.a-star.edu.sg
}

\begin{document}

\maketitle

\begin{abstract}
Deep networks are prone to performance degradation when there is a domain shift between the source (training) data and target (test) data. Recent test-time adaptation methods update batch normalization layers of pre-trained source models deployed in new target environments with streaming data to mitigate such performance degradation. Although such methods can adapt on-the-fly without first collecting a large target domain dataset, their performance is dependent on streaming conditions such as mini-batch size and class-distribution, which can be unpredictable in practice. In this work, we propose a framework for few-shot domain adaptation to address the practical challenges of data-efficient adaptation. Specifically, we propose a constrained optimization of feature normalization statistics in pre-trained source models supervised by a small support set from the target domain. Our method\footnote{Demo available at \url{https://github.com/zwenyu/lccs}.} is easy to implement and improves source model performance with as few as one sample per class for classification tasks. Extensive experiments on 5 cross-domain classification and 4 semantic segmentation datasets show that our method achieves more accurate and reliable performance than test-time adaptation, while not being constrained by streaming conditions.
\end{abstract}

\section{Introduction}
\label{sec: introduction}

While deep neural networks have demonstrated remarkable ability in representation learning, their performance relies heavily on the assumption that training \emph{(source domain)} and test \emph{(target domain)} data distributions are the same. However, as real-world data collection can be difficult, time-consuming or expensive, it may not be feasible to adequately capture all potential variation in the training set, such that test samples may be subject to \emph{domain shift} (also known as \emph{covariate shift}) due to factors such as illumination, pose and style~\cite{gulrajani2020domainbed,wilds}. To prevent severe performance degradation when models are deployed, timely adaptation to the target test distribution is needed. 

To carry out this adaptation, a range of methods with varying requirements on the availability of source and target domain data have been developed. 
In the classic \emph{domain adaptation} (DA) setting, methods assume source and target data are jointly available for training~\cite{wilson2020dasurvey}, which compromises the privacy of source domain data. To address this, methods for the \emph{source-free DA} setting~\cite{qiu2021prototypegen,yang2021generalized,liang2020shot} instead adapt a pre-trained source model using only unlabeled target data, but they still require access to the entire unlabeled target dataset like traditional DA methods. This can delay adaptation, or even make it impractical, when collecting the unlabeled target data is costly in terms of time or other resources.
%This can delay adaptation due to data collection and hence be impractical when data collection is costly in terms of time and other resources.

\begin{figure}[b]
    \centering
    \vspace{-2mm}
    \includegraphics[width=0.99\linewidth]{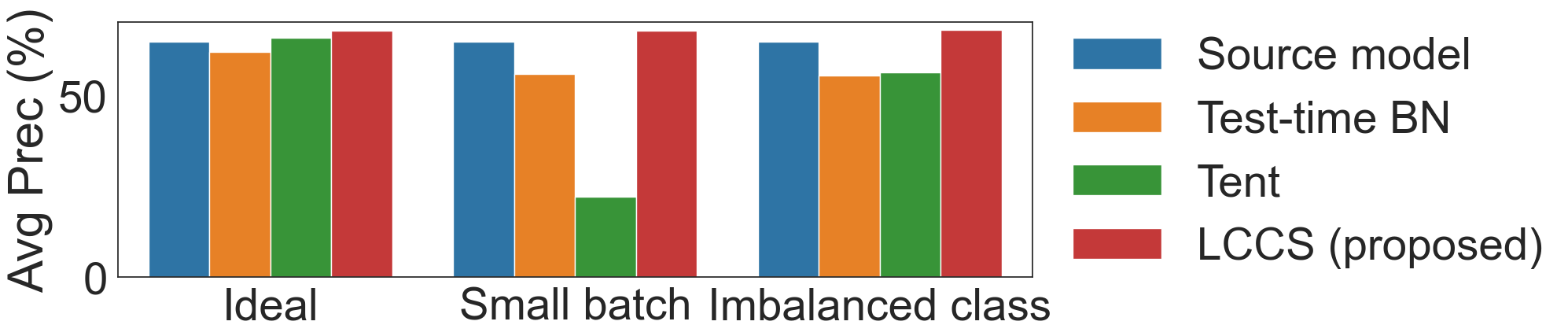}
    \vspace{-2mm}
    \caption{Comparison of cross-domain classification performance on VisDA under various test mini-batch (streaming) conditions.}
    \vspace{-2mm}
    \label{fig:overview_evaluation}
\end{figure}

\begin{table*}[h]
\centering
\begin{adjustbox}{max width=\textwidth}
\begin{tabular}{lcp{5cm}p{5cm}p{5cm}c}
\toprule[1pt]\midrule[0.3pt]
\textbf{Setups}     & \multicolumn{1}{c}{\textbf{Source-free}}          & \multicolumn{3}{c}{\textbf{Training inputs}}      & \textbf{Test inputs} \\ \cmidrule{3-5}
                    &                                                   & \multicolumn{1}{c}{Source domain(s)}              & \multicolumn{1}{c}{Target domain} & \multicolumn{1}{c}{Size of available target data} \\ \midrule
Domain adaptation   & \xmark                                            & $L^{s_1},\dots,L^{s_N}$                           & entire $U^{t}$ & $|U^{t}|$ & $U^{t}$\\
$k$-shot domain adaptation & \xmark                                     & $L^{s_1},\dots,L^{s_N}$                    & $k$-shot support set $L^{spt} \subset L^{t}$ & $k \times \text{\# classes}$ & $U^{t}$\\
Domain generalization
                    & \xmark                                            & $L^{s_1},\dots,L^{s_N}$                           & - & 0 & $U^{t}$\\ \midrule
Source-free domain adaptation   
                    & \cmark                                            & Pre-trained model on $L^{s_1},\dots,L^{s_N}$      & entire $U^{t}$ & $|U^{t}|$ & $U^{t}$\\
Test-time adaptation
                    & \cmark                                            & Pre-trained model on $L^{s_1},\dots,L^{s_N}$      & mini-batch $U^{t}$ & $|\text{mini-batch}|^*$, typically 128 & $U^{t}$\\
\textbf{Source-free $k$-shot adaptation}
                    & \cmark                                            & Pre-trained model on $L^{s_1},\dots,L^{s_N}$      & $k$-shot support set $L^{spt} \subset L^{t}$ & $k \times \text{\# classes}$ & $U^{t}$\\
\midrule[0.3pt]\bottomrule[1pt]
\end{tabular}
\end{adjustbox}
\vspace{-2mm}
\caption{Setups where $s$ and $t$ denote source and target domains. $L^d$ and $U^d$ denote labeled and unlabeled datasets from domain $d$. $*$ denotes sample size to start adaptation, and model completes adaptation after passing through entire $U^t$. $|U^t|$ can be as large as 55,000 for VisDA. \label{tab:setup}}
\vspace{-4mm}
\end{table*}
Very recently, \emph{test-time adaptation} methods~\cite{nado2020bn,schneider2020covariate,wang2021tent} have been proposed to adapt pre-trained source models on-the-fly without first collecting a large target domain dataset, by updating batch normalization (BN) layers in pre-trained source models with every mini-batch of target domain data. While these methods reduce the delay to initiate adaptation, they face three main challenges: 1) there is no guarantee that the self-supervised or unsupervised objectives used in these methods can correct domain shift without using target domain labels, 2) performance is dependent on 
having large mini-batches to obtain good estimates of BN parameters and statistics, 
%large batch size to obtain good estimates of BN parameters and statistics, 
3) test samples need to be class-balanced, which is not always the case in real-world deployments. We show in Figure~\ref{fig:overview_evaluation} on the VisDA benchmark that while in ideal conditions (large batch sizes, balanced classes) test-time adaptation methods can improve performance of source models, their performance is severely affected outside of these ideal conditions; in fact, for one of the methods, we observe catastrophic failure where all outputs “collapse” to one or a few classes. 

In this work, we propose a different solution to the higher-level objective of source-free domain adaptation with limited target domain data. We suggest that adaptation with a few labeled target domain samples instead of mini-batches of unlabeled data can address the identified challenges with test-time adaptation methods.
We propose a new method under a new \emph{source-free $k$-shot adaptation} setting, that adapts a pre-trained source model using $k$-shot labeled (or \emph{support}) samples from the target domain; a comparison with existing DA settings is shown in Table~\ref{tab:setup}. Our proposed method adapts batch normalization (BN) layers of deep source models with the support samples using a reparameterization of BN-layers and a low-dimensional approximation of the optimal target domain BN statistics. Although BN layer modulation has been explored for DA~\cite{Chang2019,nado2020bn,schneider2020covariate,wang2021tent}, reliably optimizing BN layers with extremely few support samples (as few as one per class) is a new and challenging problem. Naively optimizing high-dimensional network parameters directly risks a severely \emph{ill-posed} problem caused by data scarcity.
Our method approximates optimal target domain BN statistics with a small set of spanning vectors representing both source and support datasets and finetunes only the combination coefficients,
significantly reducing the number of parameters that need to be learned 
during adaptation. 
Our method is inspired by the success of controlling sample stylization through BN layers~\cite{huang2017adain,zhou2021mixstyle,Nam2018BatchInstanceNF}, and we aim to approximate the optimal style to stylize the target samples to address domain shift. 

More broadly, our proposed source-free $k$-shot adaptation setting addresses salient challenges associated with data availability and inference efficiency in real-world deployments. Specifically, the setting helps to protect the privacy of source domains, and has low requirements for target data availability of only $k$ labeled samples per class during adaptation. During inference, test batches can be of \emph{any} size and class composition with no restrictions. As model parameters are not further updated at test-time after $k$-shot adaptation, our proposed setting enables more efficient inference in comparison to the test-time adaptation setting.

We extensively evaluate our proposed method on 5 image classification and 4 semantic segmentation benchmark datasets and show that it achieves competitive adaptation performance through comprehensive experimental validations and comparisons with state-of-the-art methods.

\section{Related Work}
\label{sec: related_works}

\noindent\textbf{Domain adaptation (DA).}
In the DA setting, a network is trained jointly on labeled source dataset and unlabeled target dataset to optimize task performance on the target domain, as surveyed in \cite{wilson2020dasurvey}. Existing source-free DA approaches adapt a pre-trained source model with unlabeled target dataset by generating source-like representations~\cite{qiu2021prototypegen}, making use of clustering structures in target features for classification~\cite{yang2021generalized,liang2020shot}, or aligning target samples to source hypothesis through entropy minimization and information maximization~\cite{liang2020shot,Kundu2020TowardsIM,Li_2020_CVPR}. While these methods are similar to our proposed method in not accessing source data, they use a large number of unlabeled target samples for adaptation.

\noindent\textbf{Domain generalization (DG).} 
DG methods aim to learn a model robust to unseen domain shifts by training only on labeled source datasets~\cite{gulrajani2020domainbed,wilds}. Our proposed method can work directly on source models learned by DG strategies, and hence complements existing DG methods to further close the generalization gap on specific domains of interest.

\noindent\textbf{Test-time adaptation.}
Test-time adaptation methods update a pre-trained source model continuously during test time with unlabeled target samples to improve performance on the target distribution. \cite{Sun2019TestTimeTF} update network parameters guided by image rotation estimation. However, the method requires training with the self-supervised task on the source dataset, and hence does not work directly with any pre-trained source model. Recent works propose adapting BN layers of the source model instead~\cite{nado2020bn,schneider2020covariate,li2016AdaBN}. These test-time BN methods re-estimate BN statistics on each mini-batch with mini-batch evaluation to correct domain shifts due to shifts in first and second moments of the data distribution~\cite{nado2020bn,schneider2020covariate}. Tent~\cite{wang2021tent} further updates batch-norm weights and biases with entropy minimization to obtain more confident predictions. However, the reliability of test-time adaptation depends on the number and class-distribution of samples in each mini-batch.

\noindent\textbf{Few-shot transfer learning.}
There is a wide range of works that learn a metric space specifically for $k$-shot tasks, and use meta-learning to learn adaptation strategies~\cite{pan2010surveytransfer}. These approaches typically need specific network architectures, loss functions, training strategies requiring multiple source domains or joint training with source and support data together. Since they do not directly work with a pre-trained source model, we do not focus on them here.

Another popular strategy is model finetuning or weight transfer from a pre-trained source model, and our proposed method also falls in this category. Directly finetuning all source model weights on a limited support set with small $k$ is known to severely overfit, so a support set of at least $k=100$ is often required~\cite{yosinski2014howtransferable,Scott2018AdaptedDE}. Recent works constrain the dimensionality of learnable parameter space. FLUTE~\cite{triantafillou2021flute} finetunes BN parameters with a nearest-centroid classifier. \cite{Yoo2018EfficientKL} clusters network parameters and constrains all parameters in a cluster to share the same update, but this method requires activations on the source dataset for clustering.

\section{Proposed Method}
\label{sec: proposed_method}

We first revisit BN layer optimization for DA in Section~\ref{sec: bn_operation} and establish that optimizing the BN layer can be achieved by optimizing BN statistics $\{\boldsymbol{\mu},\boldsymbol{\sigma}\}$. Then, we introduce our proposed method that optimizes $\{\boldsymbol{\mu},\boldsymbol{\sigma}\}$ on a $k$-shot support set $\boldsymbol{L^{spt}}$ with a supervised loss in Section~\ref{sec: approximation_bn_layer} in the context of classification. For classification tasks, the support set contains $k$ labeled target samples $(\boldsymbol{x},\boldsymbol{y})$ per class, with image $\boldsymbol{x}$ and one-hot class vector $\boldsymbol{y}$. For semantic segmentation, the support set contains $k$ labeled target samples in total, and $\boldsymbol{y}$ is the segmentation map corresponding to image $\boldsymbol{x}$. Implementation details are described in Section~\ref{sec: implementation}. 

\subsection{Revisiting BN Layer Optimization for DA}
\label{sec: bn_operation}

The batch-norm layer operation $f$, shown below in Equation~\ref{eqn: bn}, is a function acting on input features $\mathbf{Z}$ parameterized by four variables: BN statistics $\{\boldsymbol{\mu},\boldsymbol{\sigma}\}$ and BN parameters $\{\boldsymbol{\gamma},\boldsymbol{\beta}\}$. Here $\boldsymbol{\mu}=\mathbb{E}[\mathbf{Z}] \in \mathbb{R}^C$ and $\boldsymbol{\sigma}=\sqrt{\mathbb{V}[\mathbf{Z}] + \epsilon} \in \mathbb{R}^C$ are channel-wise feature statistics for $C$ channels, and $\boldsymbol{\gamma},\boldsymbol{\beta} \in \mathbb{R}^C$ are layer weight and bias parameters. 
\begin{align}
    \mathbf{Z_{BN}} 
    &= f(\mathbf{Z}; \boldsymbol{\mu},\boldsymbol{\sigma},\boldsymbol{\gamma},\boldsymbol{\beta}) 
    = \left( \frac{\mathbf{Z}-\boldsymbol{\mu}}{\boldsymbol{\sigma}} \right) \boldsymbol{\gamma} + \boldsymbol{\beta} \label{eqn: bn}
\end{align}
  
We denote $f(\mathbf{Z}; \boldsymbol{\mu_t},\boldsymbol{\sigma_t},\boldsymbol{\gamma_t},\boldsymbol{\beta_t})$ as the BN operation with optimal target domain BN statistics and parameters $\{\boldsymbol{\mu_t},\boldsymbol{\sigma_t},\boldsymbol{\gamma_t},\boldsymbol{\beta_t}\}$. By rewriting this operation in terms of source model BN parameters $\boldsymbol{\beta_s}$ and $\boldsymbol{\gamma_s}$
\begin{align}
     f(\mathbf{Z}; \boldsymbol{\mu_t},\boldsymbol{\sigma_t},\boldsymbol{\gamma_t},\boldsymbol{\beta_t})
     &= \left( \frac{\mathbf{Z}-\boldsymbol{\mu_t}}{\boldsymbol{\sigma_t}} \right) \boldsymbol{\gamma_t} \frac{\boldsymbol{\gamma_s}}{\boldsymbol{\gamma_s}}+ \boldsymbol{\beta_t} + \boldsymbol{\beta_s} - \boldsymbol{\beta_s} \nonumber \\  
     &= \left( \frac{\mathbf{Z}-\boldsymbol{\tilde{\mu}}}{\boldsymbol{\tilde{\sigma}}} \right) \boldsymbol{\gamma_s} + \boldsymbol{\beta_s},
     \label{eqn: bn_expand}
\end{align}
we see that it is equivalent to changing BN statistics in the source model BN operation $f(\mathbf{Z}; \boldsymbol{\tilde{\mu}},\boldsymbol{\tilde{\sigma}},\boldsymbol{\gamma_s},\boldsymbol{\beta_s})$ from 
$\boldsymbol{\tilde{\sigma}}=\boldsymbol{\sigma_s}$ and $\boldsymbol{\tilde{\mu}}=\boldsymbol{\mu_s}$
to
$\boldsymbol{\tilde{\sigma}}= \frac{\boldsymbol{\sigma_t}\boldsymbol{\gamma_s}}{\boldsymbol{\gamma_t}}$ and $\boldsymbol{\tilde{\mu}} = \boldsymbol{\mu_t} - \frac{(\boldsymbol{\beta_t}-\boldsymbol{\beta_s})\boldsymbol{\sigma_t}}{\boldsymbol{\gamma_t}}$, where $\{\boldsymbol{\mu_s},\boldsymbol{\sigma_s},\boldsymbol{\gamma_s},\boldsymbol{\beta_s}\}$ are BN statistics and parameters in the pre-trained source model. 
This observation implies that we can obtain the optimal target domain BN layer $f(\mathbf{Z}; \boldsymbol{\mu_t},\boldsymbol{\sigma_t},\boldsymbol{\gamma_t},\boldsymbol{\beta_t})$ by optimizing only the BN \emph{adaptation statistics} $\{\boldsymbol{\tilde{\mu}}, \boldsymbol{\tilde{\sigma}}\}$. Edge cases of zero-valued elements in $\boldsymbol{\gamma_s}$ can be avoided by adding a small number to such elements. Setting elements as zero in $\boldsymbol{\gamma_t}$ implies that the corresponding features are completely ignored, which we assume unlikely given a well-trained source feature extractor. A detailed explanation is provided in Appendix B.

\subsection{Low Dimensional Approximation for BN Layer Adaptation}
\label{sec: approximation_bn_layer}

\noindent\textbf{Overall formulation.} In our setting with extremely few support samples, optimizing BN layers by estimating the high-dimensional BN parameters risks solving an ill-posed problem caused by data scarcity. 
We hypothesize that source and support set statistics are related to those of the target domain due to shared classes and the relatedness of domains, 
%We hypothesize that source and support set statistics at least partially represent target domain due to shared classes and the relatedness of domains, 
and propose to approximate target BN statistics in the linear span of these available statistics. We assume a linear relationship as it can be interpreted as style mixing and for computational efficiency. 
Specifically, for a BN layer with $C$ channels, we approximate the optimal target BN statistics by
%\cscomment{Can put both equations on the same line to save space}
\begin{align}
    \boldsymbol{\mu_{LCCS}} &=
    \boldsymbol{M}\boldsymbol{\eta} = 
    [\boldsymbol{\mu_s} \hspace{0.2cm} \boldsymbol{M_{spt}}] [\eta_{s} \hspace{0.2cm} \boldsymbol{\eta_{spt}}^T]^T \label{eqn: lccs_general_mu} \\ 
    \boldsymbol{\sigma_{LCCS}} &=
    \boldsymbol{\Sigma}\boldsymbol{\rho} = 
    [\boldsymbol{\sigma_s} \hspace{0.2cm} \boldsymbol{\Sigma_{spt}}] [\rho_{s} \hspace{0.2cm} \boldsymbol{\rho_{spt}}^T]^T \label{eqn: lccs_general_sigma}
\end{align}
where $\boldsymbol{M}, \boldsymbol{\Sigma}\in \mathbb{R}^{C\times (n+1)}$ contain $n+1$ spanning vectors for first and second moment statistics, and $\boldsymbol{\eta}, \boldsymbol{\rho} \in \mathbb{R}^{n+1}$ are the learnable Linear Combination Coefficients for batch normalization Statistics (LCCS). 
Our proposed formulation generalizes the original BN operation where $\boldsymbol{\eta_{spt}}=\boldsymbol{\rho_{spt}}= \boldsymbol{0}$ and $\eta_s=\rho_s=1$. It also generalizes channel-wise finetuning of BN statistics and parameters: by setting support set spanning vectors as basis vectors and $n=C$, our method is equivalent to unconstrained optimization of BN layers. The key advantage of our approach is that when $n<<C$, we can optimize BN layers using significantly fewer learnable parameters. 

\noindent\textbf{Spanning vectors.} We first describe how source and support set spanning vectors are obtained.

\emph{Source domain spanning vectors ($\boldsymbol{\mu_s}, \boldsymbol{\sigma_s}$):} We directly utilize BN statistics $\{\boldsymbol{\mu_s}, \boldsymbol{\sigma_s}\}$ from the pre-trained source model. We incorporate source representations to benefit from knowledge shared between source and target domains. 

\emph{Support set spanning vectors ($\boldsymbol{M_{spt}}, \boldsymbol{\Sigma_{spt}}$):} The support set is represented by $n$ spanning vectors extracted by aggregation functions or dimensionality reduction methods such as singular value decomposition (SVD). We choose vectors from the orthogonal basis vectors of per-sample feature statistics at the BN layer.
Specifically, for $\boldsymbol{M_{spt}}$ (and similarly for $\boldsymbol{\Sigma_{spt}}$), let $\mathbf{\tilde{Z}_\mu}\in \mathbb{R}^{C\times |\boldsymbol{L^{spt}}|}$ denote the matrix of channel-wise feature mean for each sample. We set the first spanning vector as the overall channel-wise feature mean $\boldsymbol{\mu_{spt}}$ computed across all support samples. We then subtract this first vector as $\mathbf{\tilde{Z}_\mu^{(1)}} = \mathbf{\tilde{Z}_\mu} - \mathbf{\tilde{Z}_\mu}\boldsymbol{\mu_{spt}} \boldsymbol{\mu_{spt}}^T / \|\boldsymbol{\mu_{spt}}\|^2$, factorize $\mathbf{\tilde{Z}_\mu^{(1)}}=\boldsymbol{USV^T}$ using SVD, and extract the top $n-1$ vectors in $\boldsymbol{US}$ as the remaining spanning vectors.
The resulting features post-BN are then $\mathbf{Z_{BN}} = \left( \frac{\mathbf{Z}-\boldsymbol{\mu_{LCCS}}}{\boldsymbol{\sigma_{LCCS}}} \right) \boldsymbol{\gamma_s} + \boldsymbol{\beta_s}$.

\begin{table}
\centering
\begin{adjustbox}{max width=0.9\linewidth}
\begin{tabular}{lrrr}
\toprule[1pt]\midrule[0.3pt]
\textbf{Network}    & \textbf{\# parameters} & \textbf{\# BN parameters} &\textbf{\# LCCS}  \\ \midrule
ResNet-18 & 12 million & 9,600 & 80 \\
ResNet-50 & 26 million & 53,120 & 212 \\
ResNet-101 & 45 million & 105,344 & 416 \\
DenseNet-121 & 29 million & 83,648 & 484 \\
\midrule[0.3pt]\bottomrule[1pt]
\end{tabular}
\end{adjustbox}
\vspace{-2mm}
\caption{Comparison of parameter counts for different networks. \label{tab:bn_layers}}
\vspace{-5mm}
\end{table}

\noindent\textbf{Learning LCCS parameters.} We finetune LCCS parameters $\{\boldsymbol{\eta}, \boldsymbol{\rho}\}$ from all BN layers by minimizing cross-entropy on the support set. After finetuning the LCCS parameters, we use the estimated BN adaptation statistics $\boldsymbol{\mu_{LCCS}}$ and $\boldsymbol{\sigma_{LCCS}}$ on the target domain for inference. Note that we do not adjust the BN statistics further during test time. Hence, unlike test-time BN, our adapted model is not affected by test-time mini-batch size and class distribution.

\noindent\textbf{Comparison of parameter counts.} In Table~\ref{tab:bn_layers}, we summarize the number of network parameters versus LCCS parameters when $n=1$ for common network architectures.  Model finetuning involves updating network parameters according to a new objective. Without sufficient labeled samples, updating a large number of parameters is an ill-posed problem. While adapting only the BN parameters allows a smaller number of parameters to be tuned,
%smaller adjustments to the original network, 
this is still a large number of learnable parameters in deep networks. Our proposed method dramatically reduces this number to only 4 LCCS parameters per BN layer when $n=1$, a greater than 100-fold reduction compared to the number of BN parameters. As a result, even when provided with only an extremely small support set, our proposed method is less prone to overfitting.

\subsection{Implementation Details}
\label{sec: implementation}

The overall adaptation objective is to finetune LCCS parameters to minimize cross-entropy loss $\mathcal{L}(\boldsymbol{\eta}, \boldsymbol{\rho}) = -\sum_{\boldsymbol{(x, y)}\in \boldsymbol{L^{spt}}} \boldsymbol{y} \log h(\boldsymbol{x}; \boldsymbol{\eta}, \boldsymbol{\rho})$ on the support set,
where $\boldsymbol{x}$ and $\boldsymbol{y}$ are input and one-hot class encoding of support samples $\boldsymbol{L^{spt}}$, and $h$ is the source model with learnable LCCS parameters $\{\boldsymbol{\eta},\boldsymbol{\rho}\}$. The proposed method comprises an initialization stage and a gradient update stage. 

%\cscomment{Can the use of EMA be motivated?}
\noindent \textbf{Initialization stage.} We search for initialization values for LCCS to warm start the optimization process. We first compute the support BN statistics $\boldsymbol{\mu_{spt}}$ and $\boldsymbol{\sigma_{spt}}$ by exponential moving average (EMA) for $m$ epochs to allow $\boldsymbol{\mu_{spt}}$ and $\boldsymbol{\sigma_{spt}}$ to update smoothly. Then, we conduct a one-dimensional grid search on the LCCS parameters by setting 
$\boldsymbol{\eta_{spt}}=\boldsymbol{\rho_{spt}}=[v \: 0 \cdots 0]^T$ where $v \in \{0,0.1,\dots,1.0\}$ and $\eta_s=\rho_s=1-v$ 
with values tied across all BN layers. The initialization value that minimizes cross-entropy loss on the support samples is selected.

\noindent \textbf{Gradient update stage.} We compute support set spanning vectors $\boldsymbol{M_{spt}}$ and $\boldsymbol{\Sigma_{spt}}$ with initialized LCCS parameters, and further optimize the parameters with gradient descent for $m$ epochs. In this stage, parameter values are not tied across BN layers and we do not impose sum-to-one constraints on the coefficients to allow more diverse combinations.

We set $m=10$ epochs in all our experiments. Support samples are augmented with the same data augmentations for source model training. We retain the pre-trained source classifier with extremely small support sets, and update it when sufficient support samples are available to represent the target domain and learn a new target classifier. In our experiments, we use the nearest-centroid (NCC) classifier following \cite{triantafillou2021flute} as the default classifier for $k\geq 5$.
\section{Experiments}
\label{sec: experiments_and_results}

We evaluate on image classification and segmentation tasks using publicly available benchmark datasets and compare our proposed method to existing source-free methods. For each dataset, our source models are the base models trained on source domain(s) using empirical risk minimization (ERM) or DG methods that have state-of-the-art performance on that dataset.
%demonstrating state-of-the-art performance on the corresponding datasets. 
Support samples are randomly selected and experiment results are averaged over at least 3 seeds. We use a mini-batch size of 32 for classification and 1 for segmentation, and use the Adam optimizer with 0.001 learning rate for finetuning LCCS. We provide brief descriptions of task-specific experimental setups in the respective sections; detailed descriptions of the datasets, their corresponding source models and implementation are provided in Appendix C.1.

\subsection{Image Classification}
We evaluate on 5 classification datasets covering various types of shifts: PACS (7 classes in 4 domains) for style shift, VisDA (12 classes in 2 domains) for synthetic-to-real shift, Camelyon17 (2 classes in 2 domains) and iWildCam (182 classes in 2 domains) for naturally-occurring shifts, and Office (31 classes in 3 domains) for objects photographed in different environments. We adopt evaluation metrics used by recent works: accuracy for PACS and Camelyon17, macro-F1 for iWildCam, average precision for VisDA and average per-class accuracy for Office. For LCCS we set $n=1$ on iWildCam due to memory constraints, and $n=k\times \text{\# classes}$ otherwise. 

\noindent\textbf{Comparison to test-time adaptation.}
\begin{table*}[ht]
\centering
\begin{adjustbox}{max width=0.8\textwidth}
\begin{tabular}{p{3cm}P{0.8cm}P{0.9cm}P{0.8cm}P{0.8cm}P{0.8cm}P{0.9cm}P{0.8cm}P{0.8cm}P{1.6cm}P{1.6cm}P{1.6cm}}
\toprule[1pt]\midrule[0.3pt]
\textbf{Method} & \multicolumn{4}{c}{\textbf{PACS; ERM}} & \multicolumn{4}{c}{\textbf{PACS; MixStyle}} & \textbf{Camelyon17} & \textbf{iWildCam} & \textbf{VisDA} \\ \cmidrule(lr){2-5} \cmidrule(lr){6-9} \cmidrule(lr){10-12}
    & \textbf{Art}  & \textbf{Cartoon}  & \textbf{Photo}    & \textbf{Sketch} 
    & \textbf{Art}  & \textbf{Cartoon}  & \textbf{Photo}    & \textbf{Sketch}  \\ \midrule 
Source model 
    & 76.4              & 75.8              & 96.0              & 67.0
    & 83.9              & 79.1              & 95.8              & 73.8
    & 70.3              & \underline{31.0}  & 64.7 \\
+ Test-time BN
    & 81.0              & 79.8              & 96.2              & 67.5
    & 83.3              & 82.1              & 96.7              & 74.9
    & 89.9              & 30.5              & 60.7  \\
+ Tent (Adam)
    & 83.5              & 81.8              & 96.8  & 71.3
    & \underline{86.0}  & 83.6              & 96.8              & 79.2
    & 64.1              & 18.3              & 26.5 \\
+ Tent (SGD) 
    & 81.1              & 79.6              & 96.5              & 68.2 
    & 83.7              & 82.0              & 96.4              & 75.6
    & \textbf{91.4}     & 29.9              & 65.7 \\ \midrule
+ LCCS ($k=1$) 
    & 77.9              & 80.0              & 95.9              & 72.5
    & 82.2              & 80.4              & 95.9              & 78.9
    & 76.6              & \textbf{31.8}     & 67.8 \\
+ LCCS ($k=5$) 
    & \underline{85.1}  & \underline{84.2}  & \underline{97.5}  & \underline{76.7}
    & 85.7              & \underline{85.5}  & \underline{97.2}  & \underline{80.0}
    & 88.3              & -                 & \underline{76.0} \\
+ LCCS ($k=10$)
    & \textbf{86.8}     & \textbf{86.4}     & \textbf{97.7}     & \textbf{79.4} 
    & \textbf{87.7}     & \textbf{86.9}     & \textbf{97.5}     & \textbf{83.0}
    & \underline{90.2}  & -                 & \textbf{79.2}  \\
\midrule[0.3pt]\bottomrule[1pt]
\end{tabular}
\end{adjustbox}
\vspace{-2mm}
\caption{Comparison with test-time adaptation on classification tasks: 7-class classification on PACS, binary classification on Camelyon17, 182-class classification on iWildCam, and 12-class classification on VisDA. \label{tab:results_comparison_test_time_adaptation}}
\vspace{-4mm}
\end{table*}

\begin{table}[htbp]
\centering
\begin{adjustbox}{max width=\linewidth}
\begin{tabular}{l*{7}{c}}
\toprule[1pt]\midrule[0.3pt]
\textbf{Method} &   & \multicolumn{3}{c}{\textbf{PACS}} & \multicolumn{3}{c}{\textbf{VisDA}} \\ \cmidrule(lr){3-5} \cmidrule(lr){6-8}
                        & Test batch    & Bal.  & $\alpha=10$ & $\alpha=100$    & Bal.  & $\alpha=10$ & $\alpha=100$\\ \midrule 
Source model            & any           & 83.1      & 79.9      & 76.9          & 64.7      & 55.6      & 54.7 \\
+ Test-time BN          & 8             & 78.6      & 74.6      & 63.9          & 55.7      & 52.5      & 51.5\\
                        & 32            & 83.3      & 78.1      & 66.8          & 60.7      & 57.1      & 55.2 \\
                        & 128           & 84.3      & 78.6      & 66.6          & 61.9      & 58.1      & 55.3\\
+ Tent                  & 8             & 81.1      & 77.2      & 67.8          & 22.0      & 38.7      & 46.7 \\
                        & 32            & 86.1      & 80.6      & 69.4          & 59.0      & 60.8      & 57.5 \\
                        & 128           & 86.4      & 80.0      & 67.7          & 65.7      & 59.8      & 56.2 \\ \midrule
+ LCCS ($k=1$) & any           & 84.4      & 81.3      & 78.4          & 67.8      & 67.7      & 68.0 \\
+ LCCS ($k=5$) & any           & \underline{87.1}      &\underline{84.9}      & \underline{81.9}          & \underline{76.0}      & \underline{77.2}      & \underline{77.8} \\
+ LCCS ($k=10$)& any           & \textbf{88.8}      & \textbf{86.8}      & \textbf{84.3}          & \textbf{79.2}      &\textbf{78.8}      & \textbf{79.1} \\
\midrule[0.3pt]\bottomrule[1pt]
\end{tabular}
\end{adjustbox}
\vspace{-2mm}
\caption{Classification performance in streaming settings where test-time adaptation can degrade performance: small batch-sizes and imbalanced class distribution. $\alpha$ is the ratio of samples in the largest to smallest class, and Bal. denotes a balanced class-distribution. \label{tab:tta_select_failure_cases}}
\vspace{-4mm}
\end{table}

We compare our method with the baseline source models and test-time adaptation methods that also augment only batch normalization layers. These are Test-time BN~\cite{nado2020bn} and the state-of-the-art Tent~\cite{wang2021tent} that adapts BN parameters by entropy minimization with default hyperparameters: Adam or SGD with momentum with mini-batch size 128 (larger than support set size with $k=10$ on datasets evaluated) and learning rate 0.001. 

We see from Table~\ref{tab:results_comparison_test_time_adaptation} that our proposed method improves over source models even with a single example per class for all datasets evaluated, and outperforms Test-time BN and Tent on most cases when $k\geq 5$. We observe that Tent performance is dependent on optimizer choice, with Adam outperforming SGD by approximately $2\%$ on PACS and SGD outperforming Adam by as much as $39.2\%$ on VisDA. Since the better performing optimizer is dataset-dependent, this makes the practical usage of Tent challenging.
On PACS, our proposed method outperforms the best case of Tent when the support set has 5 or 10 samples per class. 
On Camelyon17, all BN adaptation methods improve over source model accuracy by at least $18\%$ in the best case, implying that replacing source BN statistics with target BN statistics is effective in addressing domain shift in this dataset. 
On the 182-class iWildCam dataset, the test dataset is imbalanced: 102 classes have at least 1 sample, and only 87 classes have at least 5 samples. Hence, to prevent introducing more class imbalance in the adaptation and evaluation processes, we only evaluate the setting where $k=1$. Our proposed method improves over the source model, while all test-time adaptation methods degrade performance.
Similarly, on VisDA, our proposed method obtains the best performance on the target domain.

\emph{Effect of test batches:}
Although test-time adaptation methods can improve the source model without the supervision of labeled samples, their performance relies on streaming conditions and severely degrades with smaller mini-batch size and class-imbalanced mini-batches as shown in Table~\ref{tab:tta_select_failure_cases}, and in detail in Appendix C.2. We constructed long-tailed imbalanced PACS and VisDA following the procedure in \cite{cao2019imbalance}, where sample sizes decay exponentially from the first to last class with $\alpha$ being the ratio of samples in the largest to smallest class. The good performance of test-time methods is dependent on having large, class-balanced mini-batches whereas our method is independent of streaming conditions and can more reliably adapt to the target domain.

\noindent\textbf{Comparison to few-shot transfer learning.}
As far as we know, there are no existing methods designed and evaluated specifically for our source-free few-shot DA setting. We thus apply existing methods in DA and few-shot transfer learning to provide benchmark performance.
We compare with AdaBN~\cite{li2016AdaBN} by replacing source BN statistics with those calculated on target support set, finetuning the source model on BN parameters, classifier, or feature extractor, finetuning the entire model with $L^2$ or $L^2$-SP~\cite{li2018explicitTL} or DELTA~\cite{Li2019DELTADL} regularization, Late Fusion~\cite{hoffman2013oneshot} which averages scores from source and target classifier, and FLUTE~\cite{triantafillou2021flute} which optimizes BN parameters with nearest-centroid classifier. FLUTE assumes the availability of multiple source datasets to train multiple sets of BN parameters for further blending to initialize the finetuning process. Since we only have access to the pre-trained source model in our setting, we reduce FLUTE to the single source dataset case and initialize FLUTE with single source BN parameters. We follow learning settings in \cite{Li2019DELTADL} for regularized finetuning and use the SGD optimizer with momentum and weight decay $0.0004$, and $L^2$ regularization is also added to finetuning on classifier or feature extractor to prevent overfitting. For all other methods that finetune on BN layers, we use the Adam optimizer following \cite{wang2021tent}. We set learning rate 0.001, mini-batch size 32 and epochs 10 for all methods and datasets evaluated.

From Table~\ref{tab:results_source_free_few_shot}, we observe that regularized finetuning tends to adapt well at $k=1$, but performance can lag behind with larger support sets. AdaBN does not consistently improve adaptation; we show in Appendix D that completely replacing source BN statistics degrades performance for VisDA. Overall, our proposed method has the best performance in most cases.
We also compare our method to $L^2$ and FLUTE on Office in Table~\ref{tab:office31_results_resnet50}, and our method consistently outperforms on all 6 source-target combinations. Though the transfer learning methods produce comparable performance on the relatively easy dataset Camelyon17 (two classes, two domains), LCCS outperforms on more difficult datasets, which demonstrates the effectiveness of the proposed low-dimensional finetuning strategy.

\noindent\textbf{Comparison to source-free unsupervised domain adaptation.}
%\label{sec: comparison_uda}
We additionally compare our proposed method with source-free unsupervised DA methods which adapt with the entire unlabeled target dataset, including AdaBN~\cite{li2016AdaBN}, SHOT~\cite{liang2020shot}, SFDA~\cite{sfda} and SDDA~\cite{Kurmi2021sdda} in Table~\ref{tab:office31_results_resnet50} on all 6 source-target pairs in Office. We observe that self-supervision over the entire unlabeled target can produce good adaptation performance. 
However, despite using only 5 samples per class, our proposed method with finetuned linear classifier (source classifier finetuned for 200 epochs) has best adaptation performance in 5 out of 6 domain pairs, with an average accuracy of $88.9\%$ compared to $88.5\%$ obtained by state-of-the-art source-free unsupervised DA method SHOT. 

SHOT outperforms on a more challenging OfficeHome dataset (71.8\% vs 67.8\%), reflecting the difficulty of adaptation with limited data. Nonetheless, our proposed method outperforms other few-shot methods evaluated by at least $4.5\%$ on average, which demonstrates its effectiveness in the few-shot setting. Full results are provided in Appendix C.4.
\begin{table}[htb]
\centering
\begin{adjustbox}{max width=\linewidth}
\begin{tabular}{*{10}{l}}
\toprule[1pt]\midrule[0.3pt]
\textbf{Method}             & \multicolumn{3}{c}{\textbf{PACS}}  & \multicolumn{3}{c}{\textbf{Camelyon17}}  & \multicolumn{3}{c}{\textbf{VisDA}} \\ \cmidrule{2-10}
\multicolumn{1}{r}{$k=$}    & \multicolumn{1}{c}{1} & \multicolumn{1}{c}{5} & \multicolumn{1}{c}{10}
                            & \multicolumn{1}{c}{1} & \multicolumn{1}{c}{5} & \multicolumn{1}{c}{10}
                            & \multicolumn{1}{c}{1} & \multicolumn{1}{c}{5} & \multicolumn{1}{c}{10} \\ \cmidrule(lr){2-4} \cmidrule(lr){5-7} \cmidrule(lr){8-10}
AdaBN                       & 82.9 & 85.5 & 85.8            & 72.9 & 87.8 & \underline{90.2}                & 56.5 & 60.9 & 61.8 \\
finetune BN                 & 79.0 & 84.3 & 85.4            & 72.6 & 87.7 & 90.1                & 59.1 & \underline{70.9} & 74.9 \\
finetune classifier         & 82.5 & 83.7 & 83.8            & 70.5 & 70.4 & 70.5                & \underline{67.6} & 69.7 & \underline{77.4} \\
finetune feat. extractor    & \underline{83.6} & \underline{86.0} & 86.1            & \underline{79.3} & 86.5 & 88.3                & 67.3 & 68.4 & 74.7 \\
$L^2$                       & \textbf{84.4} & 85.8 & 85.6            & \textbf{79.6} & \underline{88.2} & 89.5                & 66.0 & 66.4 & 69.6 \\
$L^2$-SP                    & \textbf{84.4} & 85.8 & 85.6            & \textbf{79.6} & \underline{88.2} & 89.5                & 66.0 & 66.4 & 69.6 \\ 
DELTA                       & \textbf{84.4} & 85.8 & 85.6            & \textbf{79.6} & \underline{88.2} & 89.5                & 65.9 & 66.5 & 70.1 \\
Late Fusion                 & 83.2 & 83.6 & 83.6            & 70.4 & 70.4 & 70.5                & 67.2 & 69.8 & 74.5 \\
FLUTE                       & 73.4 & 85.8 & \underline{88.1}            & 73.1 & 86.5 & \textbf{90.9}                & 48.3 & 67.1 & 65.7 \\ \midrule
LCCS          & \textbf{84.4} & \textbf{87.1} & \textbf{88.8}            & 76.6 & \textbf{88.3} & \underline{90.2}                & \textbf{67.8} & \textbf{76.0} & \textbf{79.2} \\
\midrule[0.3pt]\bottomrule[1pt]
\end{tabular}
\end{adjustbox}
\vspace{-2mm}
\caption{Comparison with few-shot transfer learning. \label{tab:results_source_free_few_shot}}
\vspace{-2mm}
\end{table}

\begin{table}[htb]
\centering
\begin{adjustbox}{max width=\columnwidth}
\begin{tabular}{l*{8}{c}}
\toprule[1pt]\midrule[0.3pt]
\textbf{Method}         & \multicolumn{1}{c}{$\boldsymbol{k}$}  & \multicolumn{7}{c}{\textbf{Office}} \\ \cmidrule(lr){3-9}
                        &                       & $\mathbf{A\rightarrow W}$ & $\mathbf{A\rightarrow D}$ & $\mathbf{W\rightarrow A}$  
                                                & $\mathbf{W\rightarrow D}$ & $\mathbf{D\rightarrow A}$ & $\mathbf{D\rightarrow W}$ &\textbf{Avg} \\ \midrule
SHOT                    & all$^\dagger$         & 90.1                  & \textbf{94.0}                  & \underline{74.3}
                                                & \textbf{99.9}                  & \underline{74.7}                  & \underline{98.4}  & \underline{88.5} \\
SFDA                    & all$^\dagger$         & \underline{91.1}                  & \underline{92.2}                  & 71.2
                                                & 99.5                  & 71.0                  & 98.2   &  87.2 \\
SDDA                    & all$^\dagger$         & 82.5                  & 85.3                  & 67.7
                                                & \underline{99.8}                  & 71.0                  & 98.2  & 84.1 \\
AdaBN                   & all$^\dagger$         & 78.2                  & 81.3                  & 59.0
                                                & \textbf{99.9}                  & 60.3                  & 97.9  & 79.4 \\ 
L$^2$                   & 5                     & 78.9                  & 79.4                  & 64.3
                                                & \textbf{99.9}                  & 64.8                  & 97.8  & 80.9 \\
FLUTE                   & 5                     & 84.6                  & 88.2                  & 66.4
                                                & 99.1                  & 66.4                  & 95.3  & 83.3 \\ \midrule
LCCS$^*$      & 5                     & \textbf{92.8}                  & 91.8                  & \textbf{75.1}
                                                & \textbf{99.9}                  & \textbf{75.4}                  & \textbf{98.5}  &  \textbf{88.9} \\
\midrule[0.3pt]\bottomrule[1pt]
\end{tabular}
\end{adjustbox}
\vspace{-2mm}
\caption{Classification accuracy for 31-class classification on Office. Our proposed method can outperform unsupervised DA at improved data efficiency. $\dagger$ denotes target samples are unlabeled. $*$ denotes linear classifier is finetuned on support set.  \label{tab:office31_results_resnet50}}
\vspace{-3mm}
\end{table}
\begin{table}[htb]
\centering
\vspace{-2mm}
\begin{adjustbox}{max width=0.8\columnwidth}
\begin{tabular}{l*{4}c}
\toprule[1pt]\midrule[0.3pt]
\textbf{Method}     & \textbf{Cityscapes}   & \textbf{BDD-100K}     & \textbf{Mapillary}    & \textbf{SYNTHIA} \\ \midrule
ERM                 & 29.0                  & 25.1                  & 28.2                  & 26.2 \\
SW                  & 29.9                  & 27.5                  & 29.7                  & 27.6 \\
IBN-Net             & 33.9                  & 32.3                  & 37.8                  & 27.9 \\
IterNorm            & 31.8                  & 32.7                  & 33.9                  & 27.1 \\
ISW                 & 36.6                  & \underline{35.2}      & 40.3                  & \underline{28.3} \\
ISW + L$^2$         & \underline{39.5}      & 35.1                  & \underline{40.9}      & 28.1 \\ \midrule
ISW + LCCS          & \textbf{43.6}         & \textbf{37.4}         & \textbf{42.7}         & \textbf{29.1} \\
\midrule[0.3pt]\bottomrule[1pt]
\end{tabular}
\end{adjustbox}
\vspace{-2mm}
\caption{Semantic segmentation mIoU, with GTAV source domain. \label{tab:segmentation_results}}
\vspace{-6mm}
\end{table}

\begin{figure}[htb]
    \centering
    \subfigure[Target image]{\includegraphics[width=0.45\linewidth]{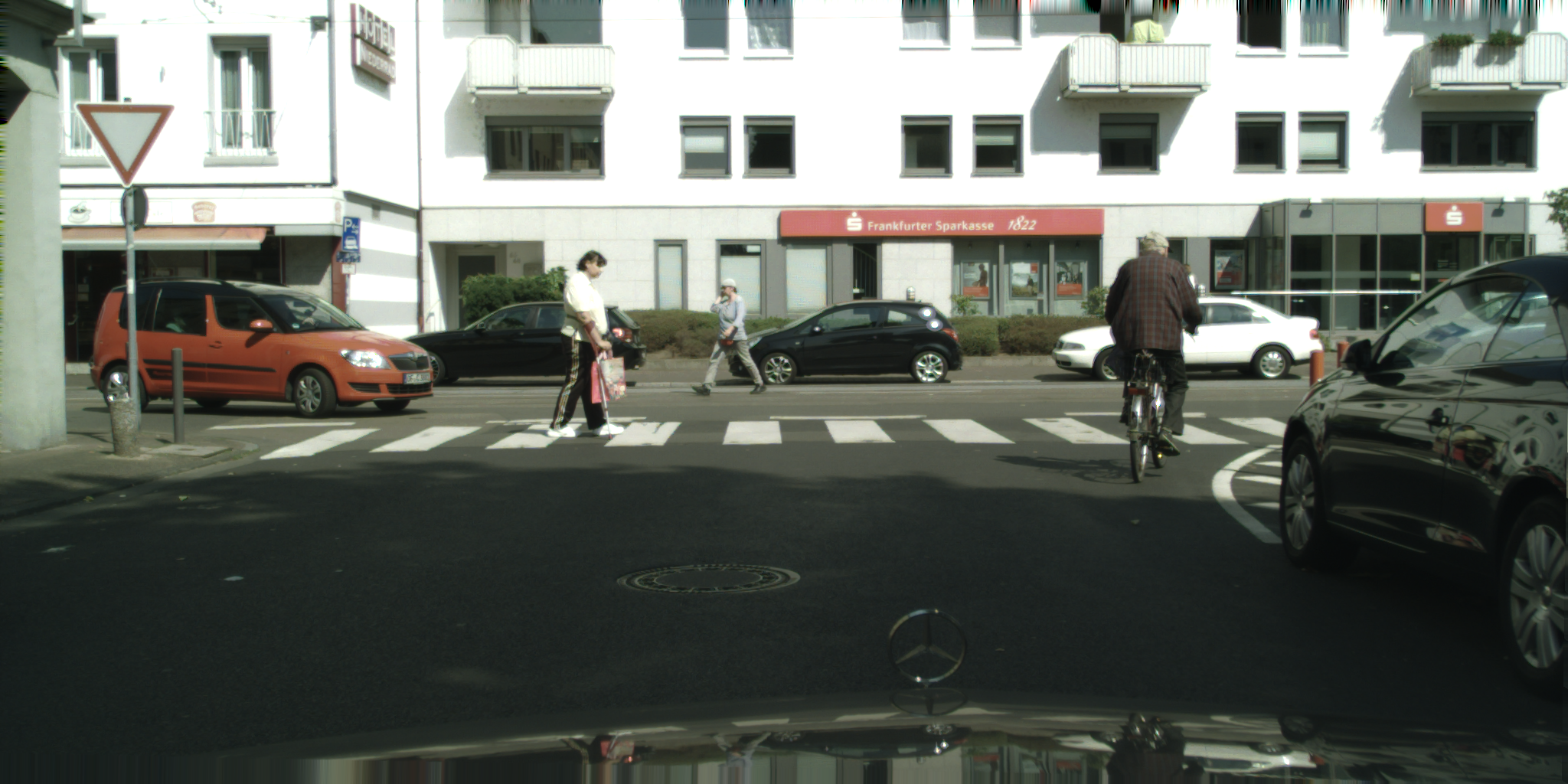}}
    \subfigure[Ground truth]{\includegraphics[width=0.45\linewidth]{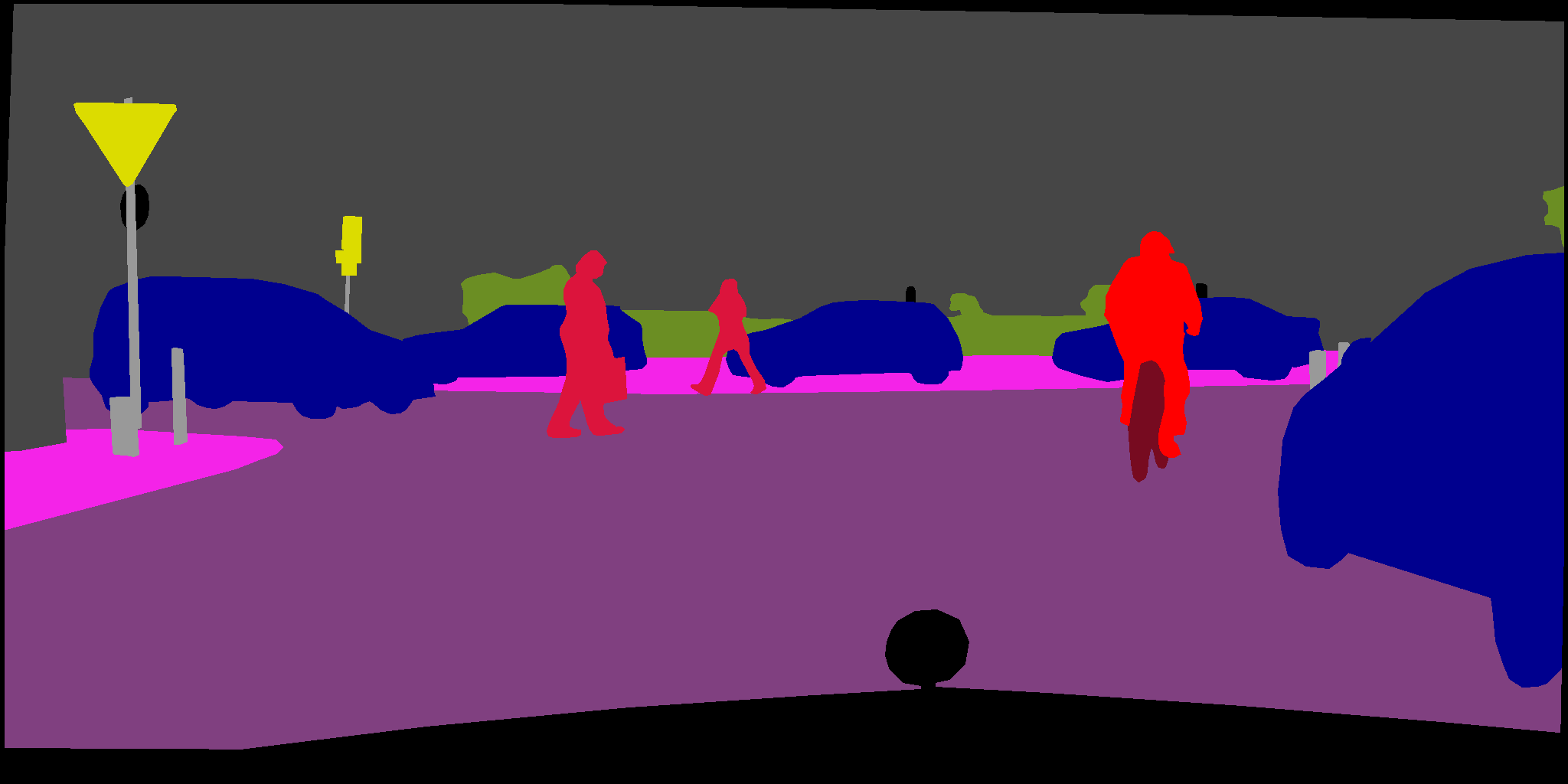}}
    \vspace{-3mm}
    
    \subfigure[ISW (SOTA)]{\includegraphics[width=0.45\linewidth]{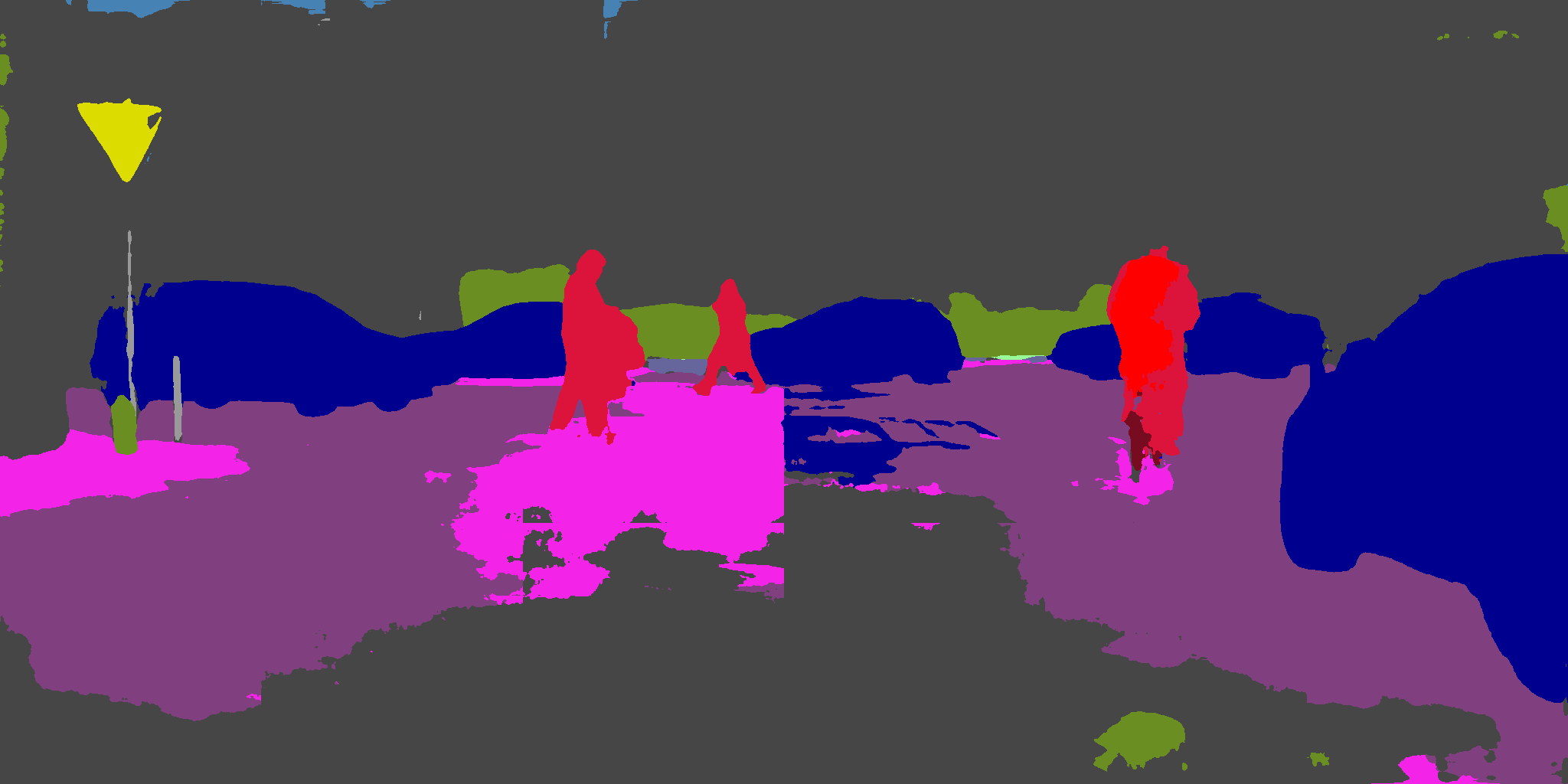}}
    \subfigure[ISW + LCCS]{\includegraphics[width=0.45\linewidth]{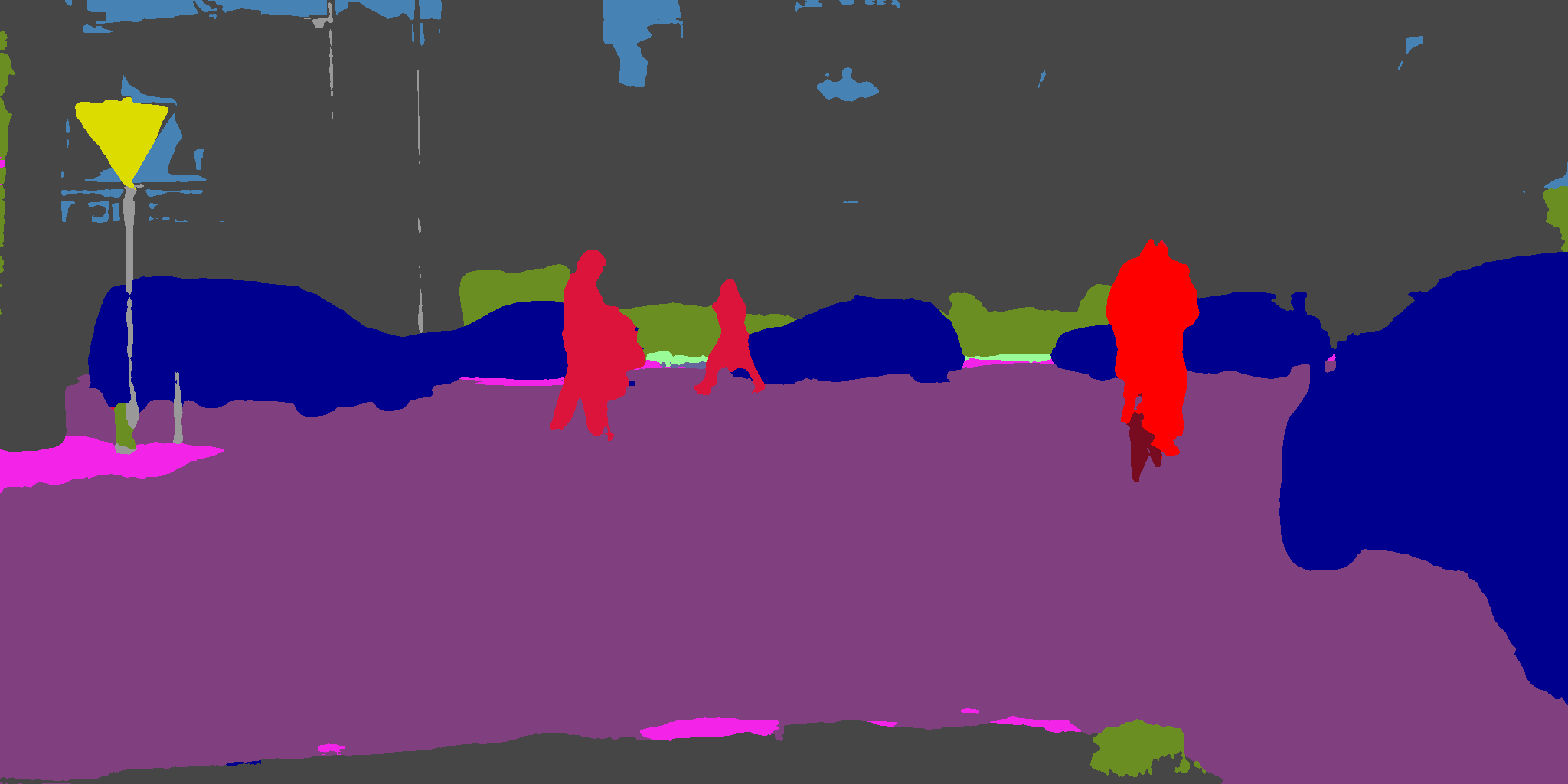}}
    \vspace{-3mm}  
    
    \caption{Semantic segmentation results for GTAV$\rightarrow$ Cityscapes.}
    \label{fig:segmentation_cityscapes}
    \vspace{-3mm}
\end{figure}

\subsection{Semantic Segmentation}
\label{sec: semantic_segmentation}

For semantic segmentation, we transfer from source dataset GTAV (synthetic) to 4 target datasets: SYNTHIA (photo-realistic), Cityscapes (real, urban), BDD-100K (real, urban-driving), and Mapillary (real, street-view). We use the mean-Intersection-over-Union (mIoU) metric to evaluate performance, and set $n=1$ for LCCS due to memory constraints. We use a total of 5 labeled images as the support set. We compare our method with SW~\cite{pan2018switchable}, IBN-Net~\cite{pan2018IBN-Net}, IterNorm~\cite{huang2019iternorms} and ISW~\cite{choi2021isw} which adapt on each sample by instance whitening or standardization, as well as $L^2$-regularized finetuning on ISW pre-trained models. From Table~\ref{tab:segmentation_results}, LCCS finetuning further improves over the state-of-the-art ISW and outperforms ISW + $L^2$ on all 4 target datasets. The improvement on Cityscapes is as much as 7\%, and is visualized in Figure~\ref{fig:segmentation_cityscapes}.
\subsection{Further Analysis}

We provide further analysis of our method in the classification setting with base configuration $n=1$ and source classifier. Additional analyses are provided in Appendix D.
\begin{table}[htb]
    \centering
    \begin{adjustbox}{max width=0.75\columnwidth}
    \begin{tabular}{ccccc}
    \toprule[1pt]\midrule[0.3pt]
    \multicolumn{2}{c}{\textbf{Stage}}                  
    & \multicolumn{3}{c}{\textbf{Avg Prec ($\%$)}} \\ \cmidrule{1-5}
    Initialization  & Gradient update   & $k=1$ & $k=5$ & $k=10$ \\ \midrule
    \xmark          & \xmark            & 64.7  & 64.7  & 64.7 \\
    \cmark          & \xmark            & 65.9  & 66.6  & 66.5 \\
    \xmark          & \cmark            & 66.0  & 66.6  & 68.6 \\
    \cmark          & \cmark            & \textbf{67.0}  & \textbf{68.1}  & \textbf{69.3} \\
    \midrule[0.3pt]\bottomrule[1pt]
    \end{tabular}
    \end{adjustbox}
    \vspace{-2mm}
    \caption{Ablation of initialization and gradient update on VisDA. \label{tab:ablation}}    
    \vspace{-1mm}
\end{table}

\begin{table}[htb!]
\centering
\begin{adjustbox}{max width=0.97\columnwidth}
\setlength\tabcolsep{4pt} % default value: 6pt
\begin{tabular}{c*{9}{l}}
\toprule[1pt]\midrule[0.3pt]
$\boldsymbol{n}$   & \multicolumn{3}{c}{\textbf{PACS}}  & \multicolumn{3}{c}{\textbf{Camelyon17}}  & \multicolumn{3}{c}{\textbf{VisDA}} \\ \cmidrule{2-10}
\multicolumn{1}{r}{$k=$}    & \multicolumn{1}{c}{1} & \multicolumn{1}{c}{5} & \multicolumn{1}{c}{10}
                            & \multicolumn{1}{c}{1} & \multicolumn{1}{c}{5} & \multicolumn{1}{c}{10}
                            & \multicolumn{1}{c}{1} & \multicolumn{1}{c}{5} & \multicolumn{1}{c}{10} \\ \cmidrule(lr){2-4} \cmidrule(lr){5-7} \cmidrule(lr){8-10}
1                           & \textbf{85.0} & 86.0 & 86.3
                            & 76.1 & 87.6 & 88.7 
                            & 67.0 & 68.1 & 69.3 \\
10                          & 84.2 & \textbf{86.2} & \textbf{86.7}
                            & \textbf{76.9} & 88.4 & 88.6 
                            & 67.7 & 69.0 & 71.2 \\
$k\times$ \# classes        & 84.4 & \textbf{86.2} & \textbf{86.7}
                            & 76.6 & \textbf{88.6} & \textbf{88.9}
                            & \textbf{67.8} & \textbf{69.5} & \textbf{72.5} \\
90\% explained var     & 84.3 & \textbf{86.2} & \textbf{86.7}
                            & \textbf{76.9} & 88.0 & 88.4
                            & 67.7 & 69.0 & 71.4\\                            
\midrule[0.3pt]\bottomrule[1pt]
\end{tabular}
\end{adjustbox}
\vspace{-2mm}
\caption{Classification accuracy with $n$ support set spanning vectors. \label{tab:results_svd}}
\vspace{-4mm}
\end{table}

\noindent\textbf{Initialization and gradient update stages.}
We conduct ablation studies on the initialization and gradient update stages of the proposed method on the VisDA dataset. From Table~\ref{tab:ablation}, we see that each stage independently improves the base model's performance, showing that both stages help adaptation.

\noindent\textbf{Design choices for optimizing LCCS.} We further experiment with different algorithm design choices in optimizing LCCS parameters. Initializing LCCS with values tied across all BN layers (67.0\% avg prec) is better than greedily initializing each BN layer sequentially starting from the shallowest layer (66.1\% avg prec). In the gradient update stage, linear combination of statistics (67.0\% avg prec) in Equations~\ref{eqn: lccs_general_mu} and \ref{eqn: lccs_general_sigma} performs better than restricting to a convex combination (66.1\% avg prec).

\noindent\textbf{Choice of $\boldsymbol{n}$.}
From Table~\ref{tab:results_svd}, we see that with larger support sets, more vectors can be used to represent the target domain and performance improves with a larger number of adaptable LCCS parameters. We observe in general that setting $n=k\times \text{\# classes}$ for $k\geq 5$ obtains the best performance, while the choice at $k=1$ is dataset-dependent.

\noindent\textbf{Computational cost.}
Our method requires little extra training time compared to that of the original network, especially with small $k$. For simplicity, consider one epoch of gradient-based training. Computation time for our proposed BN layer is $\mathcal{O}(N(A+Cn))$ for constant $A$ denoting the time for backpropagating from the loss to the BN layer's output, versus $\mathcal{O}(N(A+C))$ for the original BN layer. We expect $A >> C$ for large neural networks. Empirically, on VisDA using a Tesla V100-SXM2 GPU, the average training time per epoch on a vanilla ResNet-101 is 3.18s, 3.56s and 4.32s for $k=1$, 5 and 10 respectively. With our default LCCS parameters of  $n=k\times\text{\# classes}$, average times are 3.48s, 4.43s and 6.40s.

\vspace{-2mm}
\section{Conclusion}
\label{sec: conclusion}
In this work, we proposed the source-free $k$-shot domain adaptation setting and LCCS finetuning method. LCCS uses a low-dimensional approximation of BN statistics to significantly reduce the number of adaptation parameters, enabling robust adaptation using only a limited support set.
Our method adapts source models with as few as one sample per class, is not affected by streaming conditions, and performs well across classification and semantic segmentation benchmarks. These characteristics make our proposed solution a favourable option for data-efficient adaptation, and provide a useful foundation for future work on this challenging topic.

\section*{Acknowledgments}
This research is supported by the Agency for Science, Technology and Research (A*STAR) under its AME Programmatic Funds (Grant No. A20H6b0151).

\bibliography{references}

\begin{thebibliography}{}

\bibitem[\protect\citeauthoryear{Cao \bgroup \em et al.\egroup
  }{2019}]{cao2019imbalance}
K.~Cao, C.~Wei, A.~Gaidon, N.~Arechiga, and T.~Ma.
\newblock Learning imbalanced datasets with label-distribution-aware margin
  loss.
\newblock {\em NeurIPS}, 2019.

\bibitem[\protect\citeauthoryear{Chang \bgroup \em et al.\egroup
  }{2019}]{Chang2019}
W.-G. Chang, T.~You, S.~Seo, S.~Kwak, and B.~Han.
\newblock Domain-specific batch normalization for unsupervised domain
  adaptation.
\newblock {\em CVPR}, 2019.

\bibitem[\protect\citeauthoryear{Chen \bgroup \em et al.\egroup
  }{2021a}]{chen2021contrastive}
W.~Chen, Z.~Yu, S.~D. Mello, S.~Liu, J.~M Alvarez, Z.~Wang, and A.~Anandkumar.
\newblock Contrastive syn-to-real generalization.
\newblock {\em ICLR}, 2021.

\bibitem[\protect\citeauthoryear{Chen \bgroup \em et al.\egroup
  }{2021b}]{csgrepo}
W.~Chen, Z.~Yu, S.~D. Mello, S.~Liu, J.~M Alvarez, Z.~Wang, and A.~Anandkumar.
\newblock Contrastive syn-to-real generalization: Code repository.
\newblock \url{https://github.com/NVlabs/CSG}, 2021.

\bibitem[\protect\citeauthoryear{Choi \bgroup \em et al.\egroup
  }{2021}]{choi2021isw}
S.~Choi, S.~Jung, H.~Yun, J.~Kim, S.~Kim, and J.~Choo.
\newblock Robustnet: Improving domain generalization in urban-scene
  segmentation via instance selective whitening.
\newblock {\em CVPR}, 2021.

\bibitem[\protect\citeauthoryear{Cordts \bgroup \em et al.\egroup
  }{2016}]{cordts2016cityscapes}
M.~Cordts, M.~Omran, S.~Ramos, T.~Rehfeld, M.~Enzweiler, R.~Benenson,
  U.~Franke, S.~Roth, and B.~Schiele.
\newblock The cityscapes dataset for semantic urban scene understanding.
\newblock {\em CVPR}, 2016.

\bibitem[\protect\citeauthoryear{{Gulrajani} and
  {Lopez-Paz}}{2021}]{gulrajani2020domainbed}
I.~{Gulrajani} and D.~{Lopez-Paz}.
\newblock In search of lost domain generalization.
\newblock {\em ICLR}, 2021.

\bibitem[\protect\citeauthoryear{Hoffman \bgroup \em et al.\egroup
  }{2013}]{hoffman2013oneshot}
J.~Hoffman, E.~Tzeng, J.~Donahue, Y.~Jia, K.~Saenko, and Trevor D.
\newblock One-shot adaptation of supervised deep convolutional models.
\newblock {\em ArXiv}, 2013.

\bibitem[\protect\citeauthoryear{Huang and Belongie}{2017}]{huang2017adain}
X.~Huang and S.~Belongie.
\newblock Arbitrary style transfer in real-time with adaptive instance
  normalization.
\newblock {\em ICCV}, 2017.

\bibitem[\protect\citeauthoryear{Huang \bgroup \em et al.\egroup
  }{2019}]{huang2019iternorms}
L.~Huang, Y.~Zhou, F.~Zhu, L.~Liu, and L.~Shao.
\newblock Iterative normalization: Beyond standardization towards efficient
  whitening.
\newblock {\em CVPR}, 2019.

\bibitem[\protect\citeauthoryear{Kang \bgroup \em et al.\egroup
  }{2019}]{kang2019contrastive}
G.~Kang, L.~Jiang, Y.~Yang, and A.~Hauptmann.
\newblock Contrastive adaptation network for unsupervised domain adaptation.
\newblock {\em CVPR}, 2019.

\bibitem[\protect\citeauthoryear{Kim \bgroup \em et al.\egroup }{2020}]{sfda}
Y.~Kim, S.~Hong, D.~Cho, H.~Park, and P.~Panda.
\newblock Domain adaptation without source data.
\newblock {\em ArXiv}, 2020.

\bibitem[\protect\citeauthoryear{Koh \bgroup \em et al.\egroup }{2020a}]{wilds}
P.~Koh, S.~Sagawa, H.~Marklund, S.~Xie, M.~Zhang, A.~Balsubramani, W.~Hu,
  M.~Yasunaga, R.~Phillips, S.~Beery, J.~Leskovec, A.~Kundaje, E.~Pierson,
  S.~Levine, C.~Finn, and P.~Liang.
\newblock Wilds: A benchmark of in-the-wild distribution shifts.
\newblock {\em ArXiv}, 2020.

\bibitem[\protect\citeauthoryear{Koh \bgroup \em et al.\egroup
  }{2020b}]{wildsrepo}
P.~Koh, S.~Sagawa, H.~Marklund, S.~Xie, M.~Zhang, A.~Balsubramani, W.~Hu,
  M.~Yasunaga, R.~Phillips, S.~Beery, J.~Leskovec, A.~Kundaje, E.~Pierson,
  S.~Levine, C.~Finn, and P.~Liang.
\newblock Wilds: Code repository.
\newblock
  \url{https://worksheets.codalab.org/worksheets/0x52cea64d1d3f4fa89de326b4e31aa50a},
  2020.

\bibitem[\protect\citeauthoryear{Kundu \bgroup \em et al.\egroup
  }{2020}]{Kundu2020TowardsIM}
J.~N. Kundu, N.~Venkat, R.~Ambareesh, R.~M. V., and R.~V. Babu.
\newblock Towards inheritable models for open-set domain adaptation.
\newblock {\em CVPR}, 2020.

\bibitem[\protect\citeauthoryear{Kurmi \bgroup \em et al.\egroup
  }{2021}]{Kurmi2021sdda}
V.~Kurmi, Venkatesh~K. Subramanian, and Vinay~P. Namboodiri.
\newblock Domain impression: A source data free domain adaptation method.
\newblock {\em WACV}, 2021.

\bibitem[\protect\citeauthoryear{Li \bgroup \em et al.\egroup
  }{2016}]{li2016AdaBN}
Y.~Li, N.~Wang, J.~Shi, J.~Liu, and X.~Hou.
\newblock Revisiting batch normalization for practical domain adaptation.
\newblock {\em Pattern Recognition}, 80, 03 2016.

\bibitem[\protect\citeauthoryear{Li \bgroup \em et al.\egroup
  }{2017}]{li2018pacs}
D.~Li, Y.~Yang, Y.~Song, and T.~Hospedales.
\newblock Deeper, broader and artier domain generalization.
\newblock {\em ICCV}, 2017.

\bibitem[\protect\citeauthoryear{Li \bgroup \em et al.\egroup
  }{2018}]{li2018explicitTL}
X.~Li, Y.~Grandvalet, and F.~Davoine.
\newblock Explicit inductive bias for transfer learning with convolutional
  networks.
\newblock {\em ICML}, 2018.

\bibitem[\protect\citeauthoryear{Li \bgroup \em et al.\egroup
  }{2019}]{Li2019DELTADL}
X.~Li, H.~Xiong, H.~Wang, Y.~Rao, L.~Liu, and J.~Huan.
\newblock {DELTA}: Deep learning transfer using feature map with attention for
  convolutional networks.
\newblock {\em ICLR}, 2019.

\bibitem[\protect\citeauthoryear{Li \bgroup \em et al.\egroup
  }{2020}]{Li_2020_CVPR}
R.~Li, Q.~Jiao, W.~Cao, H.-S. Wong, and S.~Wu.
\newblock Model adaptation: Unsupervised domain adaptation without source data.
\newblock {\em CVPR}, 2020.

\bibitem[\protect\citeauthoryear{Liang \bgroup \em et al.\egroup
  }{2020}]{liang2020shot}
J.~Liang, D.~Hu, and J.~Feng.
\newblock Do we really need to access the source data? source hypothesis
  transfer for unsupervised domain adaptation.
\newblock {\em ICML}, 2020.

\bibitem[\protect\citeauthoryear{Nado \bgroup \em et al.\egroup
  }{2020}]{nado2020bn}
Z.~Nado, S.~Padhy, D.~Sculley, A.~D'Amour, B.~Lakshminarayanan, and J.~Snoek.
\newblock Evaluating prediction-time batch normalization for robustness under
  covariate shift.
\newblock {\em ArXiv}, 2020.

\bibitem[\protect\citeauthoryear{Nam and Kim}{2018}]{Nam2018BatchInstanceNF}
H.~Nam and H.-E. Kim.
\newblock Batch-instance normalization for adaptively style-invariant neural
  networks.
\newblock {\em NeurIPS}, 2018.

\bibitem[\protect\citeauthoryear{Neuhold \bgroup \em et al.\egroup
  }{2017}]{neuhold2017mapillary}
G.~Neuhold, T.~Ollmann, Samuel~R. Bul{ò}, and P.~Kontschieder.
\newblock The mapillary vistas dataset for semantic understanding of street
  scenes.
\newblock {\em ICCV}, 2017.

\bibitem[\protect\citeauthoryear{Pan and Yang}{2010}]{pan2010surveytransfer}
S.~J. Pan and Q.~Yang.
\newblock A survey on transfer learning.
\newblock {\em IEEE Transactions on Knowledge and Data Engineering},
  22(10):1345--1359, 2010.

\bibitem[\protect\citeauthoryear{Pan \bgroup \em et al.\egroup
  }{2018}]{pan2018IBN-Net}
X.~Pan, P.~Luo, J.~Shi, and X.~Tang.
\newblock Two at once: Enhancing learning and generalization capacities via
  ibn-net.
\newblock {\em ECCV}, 2018.

\bibitem[\protect\citeauthoryear{Pan \bgroup \em et al.\egroup
  }{2019}]{pan2018switchable}
X.~Pan, X.~Zhan, J.~Shi, X.~Tang, and P.~Luo.
\newblock Switchable whitening for deep representation learning.
\newblock {\em ICCV}, 2019.

\bibitem[\protect\citeauthoryear{Peng \bgroup \em et al.\egroup
  }{2017}]{Peng2017VisDATV}
X.~Peng, B.~Usman, N.~Kaushik, J.~Hoffman, D.~Wang, and Kate Saenko.
\newblock Visda: The visual domain adaptation challenge.
\newblock {\em ArXiv}, abs/1710.06924, 2017.

\bibitem[\protect\citeauthoryear{Qiu \bgroup \em et al.\egroup
  }{2021}]{qiu2021prototypegen}
Z.~Qiu, Y.~Zhang, H.~Lin, S.~Niu, Y.~Liu, Q.~Du, and M.~Tan.
\newblock Source-free domain adaptation via avatar prototype generation and
  adaptation.
\newblock {\em IJCAI}, 2021.

\bibitem[\protect\citeauthoryear{Richter \bgroup \em et al.\egroup
  }{2016}]{richter2016gtav}
S.~Richter, V.~Vineet, S.~Roth, and V.~Koltun.
\newblock Playing for data: Ground truth from computer games.
\newblock {\em ECCV}, 2016.

\bibitem[\protect\citeauthoryear{Ros \bgroup \em et al.\egroup
  }{2016}]{ros2016synthia}
G.~Ros, L.~Sellart, J.~Materzynska, D.~Vazquez, and A.~M. Lopez.
\newblock The {SYNTHIA} dataset: A large collection of synthetic images for
  semantic segmentation of urban scenes.
\newblock {\em CVPR}, 2016.

\bibitem[\protect\citeauthoryear{Saenko \bgroup \em et al.\egroup
  }{2010}]{Saenko2010AdaptingVC}
K.~Saenko, B.~Kulis, M.~Fritz, and T.~Darrell.
\newblock Adapting visual category models to new domains.
\newblock {\em ECCV}, 2010.

\bibitem[\protect\citeauthoryear{Schneider \bgroup \em et al.\egroup
  }{2020}]{schneider2020covariate}
S.~Schneider, E.~Rusak, L.~Eck, O.~Bringmann, W.~Brendel, and M.~Bethge.
\newblock Improving robustness against common corruptions by covariate shift
  adaptation.
\newblock {\em NeurIPS}, 2020.

\bibitem[\protect\citeauthoryear{Scott \bgroup \em et al.\egroup
  }{2018}]{Scott2018AdaptedDE}
T.~R. Scott, K.~Ridgeway, and M.~Mozer.
\newblock Adapted deep embeddings: A synthesis of methods for k-shot inductive
  transfer learning.
\newblock {\em NeurIPS}, 2018.

\bibitem[\protect\citeauthoryear{Sun \bgroup \em et al.\egroup
  }{2020}]{Sun2019TestTimeTF}
Y.~Sun, X.~Wang, Z.~Liu, J.~Miller, A.~A. Efros, and M.~Hardt.
\newblock Test-time training for out-of-distribution generalization.
\newblock {\em ICML}, 2020.

\bibitem[\protect\citeauthoryear{Triantafillou \bgroup \em et al.\egroup
  }{2021}]{triantafillou2021flute}
E.~Triantafillou, H.~Larochelle, R.~S. Zemel, and V.~Dumoulin.
\newblock Learning a universal template for few-shot dataset generalization.
\newblock {\em ICML}, 2021.

\bibitem[\protect\citeauthoryear{Venkateswara \bgroup \em et al.\egroup
  }{2017}]{Venkateswara2017DeepHN}
Hemanth Venkateswara, Jose Eusebio, Shayok Chakraborty, and Sethuraman
  Panchanathan.
\newblock Deep hashing network for unsupervised domain adaptation.
\newblock {\em CVPR}, 2017.

\bibitem[\protect\citeauthoryear{Wang \bgroup \em et al.\egroup
  }{2021}]{wang2021tent}
D.~Wang, E.~Shelhamer, S.~Liu, B.~Olshausen, and T.~Darrell.
\newblock Tent: Fully test-time adaptation by entropy minimization.
\newblock {\em ICLR}, 2021.

\bibitem[\protect\citeauthoryear{Wilson and Cook}{2020}]{wilson2020dasurvey}
G.~Wilson and D.~J. Cook.
\newblock A survey of unsupervised deep domain adaptation.
\newblock {\em ACM Trans. Intell. Syst. Technol.}, 11(5), July 2020.

\bibitem[\protect\citeauthoryear{Yang \bgroup \em et al.\egroup
  }{2021}]{yang2021generalized}
S.~Yang, Y.~Wang, J.~Weijer, L.~Herranz, and S.~Jui.
\newblock Generalized source-free domain adaptation.
\newblock {\em ICCV}, 2021.

\bibitem[\protect\citeauthoryear{Yoo \bgroup \em et al.\egroup
  }{2018}]{Yoo2018EfficientKL}
D.~Yoo, H.~Fan, V.~N. Boddeti, and K.~M. Kitani.
\newblock Efficient k-shot learning with regularized deep networks.
\newblock {\em AAAI}, 2018.

\bibitem[\protect\citeauthoryear{Yosinski \bgroup \em et al.\egroup
  }{2014}]{yosinski2014howtransferable}
J.~Yosinski, J.~Clune, Y.~Bengio, and H.~Lipson.
\newblock How transferable are features in deep neural networks?
\newblock {\em NIPS}, 2014.

\bibitem[\protect\citeauthoryear{Yu \bgroup \em et al.\egroup
  }{2020}]{yu2020bdd100k}
F.~Yu, H.~Chen, X.~Wang, W.~Xian, Y.~Chen, F.~Liu, V.~Madhavan, and T.~Darrell.
\newblock {BDD100K}: A diverse driving dataset for heterogeneous multitask
  learning.
\newblock {\em CVPR}, 2020.

\bibitem[\protect\citeauthoryear{Zhou \bgroup \em et al.\egroup
  }{2021a}]{zhou2021mixstyle}
K.~Zhou, Y.~Yang, Y.~Qiao, and T.~Xiang.
\newblock Domain generalization with mixstyle.
\newblock {\em ICLR}, 2021.

\bibitem[\protect\citeauthoryear{Zhou \bgroup \em et al.\egroup
  }{2021b}]{mixstylerepo}
K.~Zhou, Y.~Yang, Y.~Qiao, and T.~Xiang.
\newblock Domain generalization with mixstyle: Code repository.
\newblock \url{https://github.com/KaiyangZhou/mixstyle-release}, 2021.

\end{thebibliography}
\bibliographystyle{named}

\clearpage

\appendix
\section*{Appendix}

\section{Challenges with Adaptation by Test-time BN}
\label{sec: challenges_test_time_bn}

\begin{figure}[htb!]
    \centering
    \subfigure[\centering Effect of test-time BN in correcting domain shift \label{fig:challenges_domain_shift}]{{\includegraphics[width=\linewidth]{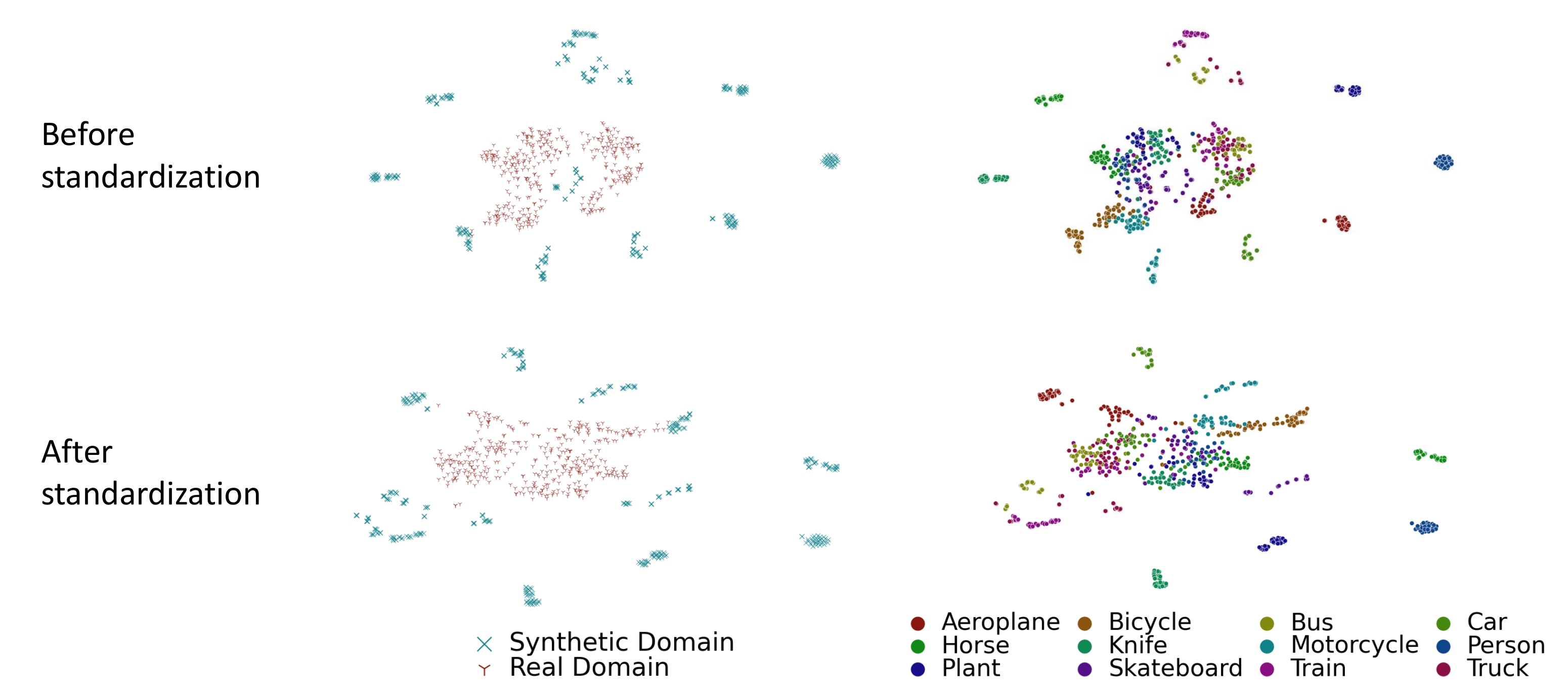}}}
    
    \subfigure[\centering Effect of sample size in estimating BN statistics; all samples from source \label{fig:challenges_sample_size}]{{\includegraphics[width=\linewidth]{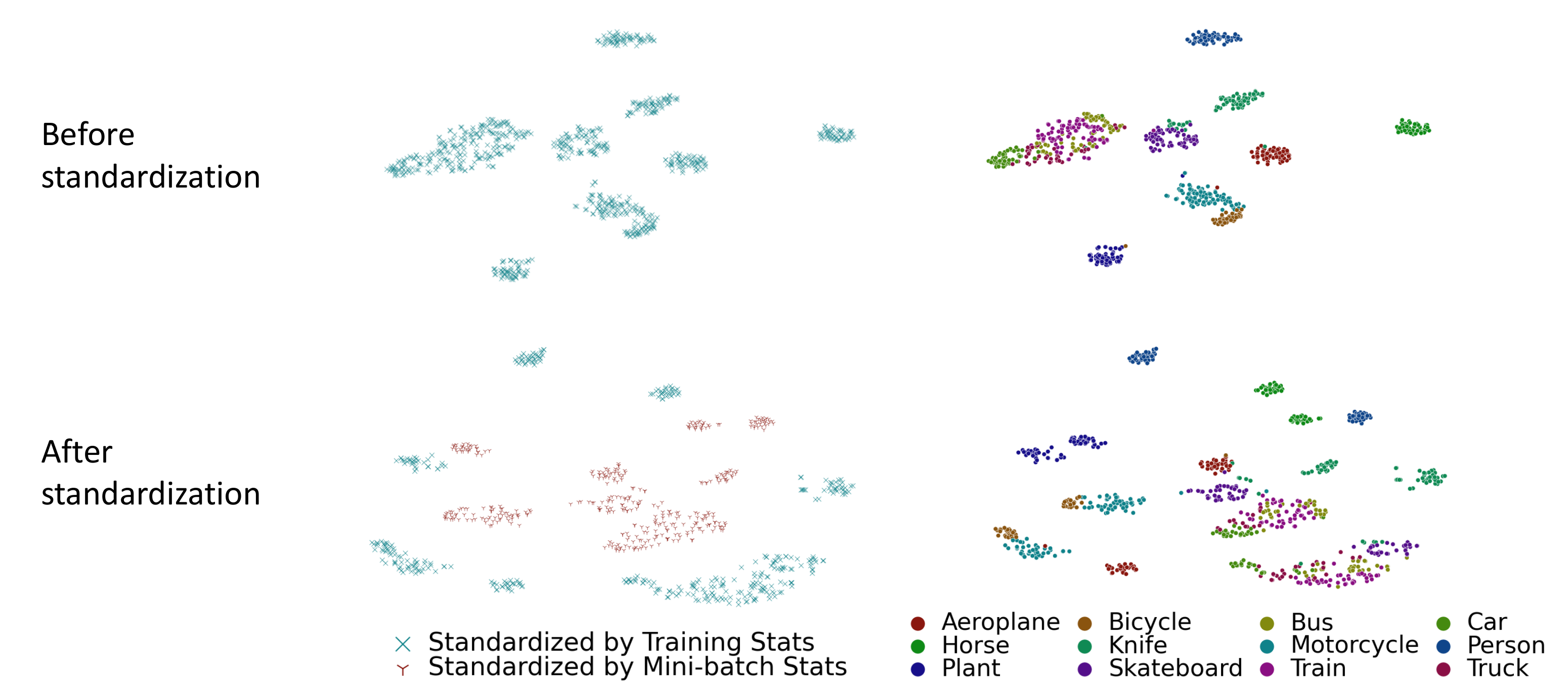}}}
    
    \subfigure[\centering Effect of class distribution in estimating BN statistics; all samples from source \label{fig:challenges_class_imbalance}]{{\includegraphics[width=\linewidth]{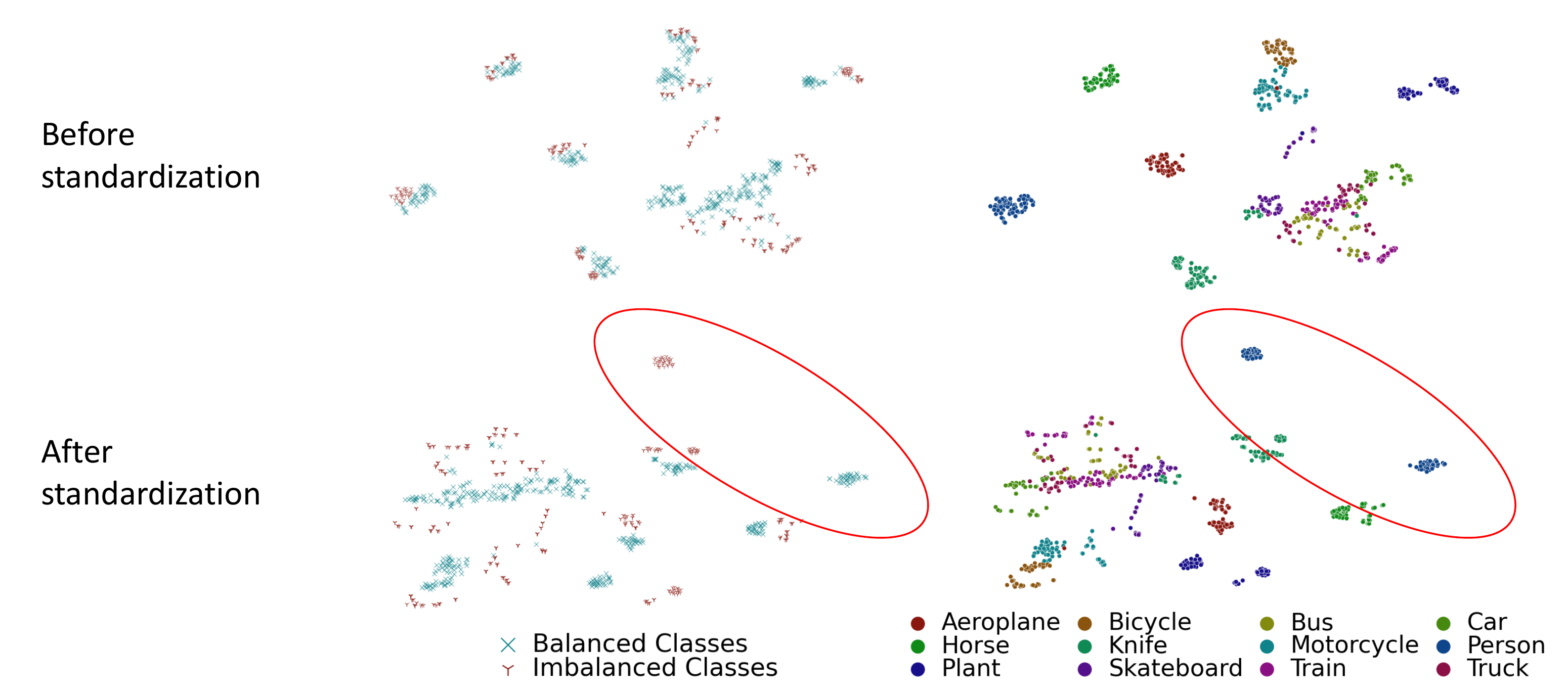}}}
    
    \caption{Challenges in applying time-time BN statistics, demonstrated on VisDA dataset with t-SNE plots of features at last BN layer of ResNet-101. (a) No guarantee that domain shift is corrected without target domain label information, (b) large mini-batch size is needed, (c) samples in each mini-batch need to be class-balanced. Best viewed in color.}
    \label{fig:challenges_test_time_bn}
\end{figure}

Empirical works in literature show that test-time BN and other methods replacing training BN statistics with test-time BN statistics can improve source model adaptation to domain shift. We identify 3 challenges when using test-time BN:
\begin{enumerate}
    \item No guarantee that domain shift can be corrected;
    \item Need for mini-batch evaluation with large mini-batch size;
    \item Need for controlled class distribution in each mini-batch;
\end{enumerate}
In Figure~\ref{fig:challenges_test_time_bn}, we illustrate these points using the source model trained on the VisDA dataset synthetic domain by the state-of-the-art method CSG~\cite{kang2019contrastive}, with mini-batch size 32. We plot t-SNE representations of features input into the last BN layer of ResNet-101, and after standardization by BN statistics.

Firstly, there is no guarantee that using mini-batch statistics on all BN layers can correct for domain shift without using any label information on the target domain. Existing works~\cite{zhou2021mixstyle} observe that sample mean and standard deviation differ across domains in the shallow layers of the network, but differ across classes in the deeper layers of the network, such that BN statistics may need to be adapted differently at different layers. In Figure~\ref{fig:challenges_domain_shift}, we see an obvious displacement of the features due to synthetic-to-real domain shift, but test-time BN does not completely map target domain features onto the source domain feature space. Moreover, entropy minimization over unlabeled data in mini-batches does not guarantee performance maximization as shown in Figure~\ref{fig:visda_diff_init}. We cannot prevent trivial solutions where all unlabeled samples are assigned the same one-hot encoding without information on the domain evaluated.

Mini-batch evaluation is required in test-time BN to estimate new BN statistics, however the quality of statistics estimated in mini-batches depends on batch size. In Figure~\ref{fig:challenges_sample_size}, we draw two sub-datasets from the source domain, so there is no domain shift. We standardize one sub-dataset with training BN statistics collected by the source model after training on the entire source dataset, and the other sub-dataset with mini-batch statistics. Although the features from the two sub-datasets match exactly before standardization, a shift is introduced from using the two different BN statistics. 

During mini-batch evaluation, each mini-batch is expected to have the same class distribution as at training. While users can enforce evenly distributed classes during training, we may not be able to control the class composition of samples at test-time when the model is deployed in the wild. To demonstrate how different class distributions in mini-batches affect the resulting features, in Figure~\ref{fig:challenges_class_imbalance}, we draw one class-balanced sub-dataset with the same number of samples per class, and a class-imbalanced sub-dataset following the original source domain class composition. For example, the original class composition has 11.4\% `motorcycle' and 4.8\% `bicycle'. Both sub-datasets are drawn from the source domain so there is no domain shift. After standardization, we observe a greater shift in the features as compared to before standardization. For instance, samples from the `person' class are split into two distinct clusters depending on the BN statistics used.

\section{BN layer parameterization}
\label{sec: appendix_bn_operation}

There are two pairs of adjustable variables in the BN operation in Equation~\ref{eqn: appendix_bn}, namely BN statistics $\{\boldsymbol{\mu},\boldsymbol{\sigma}\}$ and BN parameters $\{\boldsymbol{\gamma},\boldsymbol{\beta}\}$. Conventionally, when BN layers are finetuned, the BN parameters $\{\boldsymbol{\gamma},\boldsymbol{\beta}\}$ are optimized by gradient update on a specified task objective~\cite{wang2021tent,triantafillou2021flute}. In this section, we point out that optimizing either $\{\boldsymbol{\mu},\boldsymbol{\sigma}\}$ or $\{\boldsymbol{\gamma},\boldsymbol{\beta}\}$ is equivalent to optimizing all 4 variables except for edge cases. With this observation, we can choose to optimize BN statistics when it is more convenient to do so.

We consider the BN operation $f$ of a single BN layer for a fixed input feature $\mathbf{Z}$, and without the loss of generality, we assume the the number of channels $C=1$ in the BN layer. We define
\begin{align}
    f(\mathbf{Z}; \boldsymbol{\mu},\boldsymbol{\sigma},\boldsymbol{\gamma},\boldsymbol{\beta}) 
    = \left( \frac{\mathbf{Z}-\boldsymbol{\mu}}{\boldsymbol{\sigma}} \right) \boldsymbol{\gamma} + \boldsymbol{\beta} \label{eqn: appendix_bn}
\end{align}
where $\mathbf{Z} \in \mathbb{R}^{h \times w}$ for $h,w\in \mathbb{Z}^+$, and $\boldsymbol{\mu}, \boldsymbol{\beta} \in \mathbb{R}$, and $\boldsymbol{\sigma}, \boldsymbol{\gamma} \in \mathbb{R}\setminus\{0\}$.

Firstly, we show that adjusting $\{\boldsymbol{\gamma},\boldsymbol{\beta}\}$ is equivalent to adjusting all 4 variables i.e.
for any arbitrary
$\boldsymbol{\mu}, \boldsymbol{\tilde{\mu}}, \boldsymbol{\tilde{\beta}} \in \mathbb{R}$ and 
$\boldsymbol{\sigma}, \boldsymbol{\tilde{\sigma}}, \boldsymbol{\tilde{\gamma}} \in \mathbb{R}\setminus\{0\}$, 
there exists $\boldsymbol{\beta^*} \in \mathbb{R}$ and $\boldsymbol{\gamma^*} \in \mathbb{R} \setminus \{0\}$ such that $f(\bold{Z}; \boldsymbol{\mu},\boldsymbol{\sigma},\boldsymbol{\gamma^*},\boldsymbol{\beta^*}) = f(\bold{Z}; \boldsymbol{\tilde{\mu}},\boldsymbol{\tilde{\sigma}},\boldsymbol{\tilde{\gamma}},\boldsymbol{\tilde{\beta}})$:
By algebraic manipulation, the $\boldsymbol{\gamma^*}$ and $\boldsymbol{\beta^*}$ that fulfill the equivalence are:
\begin{align}
    \boldsymbol{\gamma^*} 
    &= \left( \frac{\boldsymbol{\sigma}}{\boldsymbol{\tilde{\sigma}}}\right) \boldsymbol{\tilde{\gamma}} 
    \hspace{0.5cm} \text{and} \hspace{0.5cm} 
    \boldsymbol{\beta^*}
    = \left( \frac{\boldsymbol{\mu}-\boldsymbol{\tilde{\mu}}}{\boldsymbol{\tilde{\sigma}}} \right) \boldsymbol{\tilde{\gamma}} + \boldsymbol{\tilde{\beta}} 
\end{align}

Next, we show that adjusting $\{\boldsymbol{\mu},\boldsymbol{\sigma}\}$ is equivalent to adjusting all 4 variables i.e.
for any arbitrary
$\boldsymbol{\tilde{\mu}}, \boldsymbol{\tilde{\beta}}, \boldsymbol{\beta} \in \mathbb{R}$ and 
$\boldsymbol{\tilde{\sigma}}, \boldsymbol{\tilde{\gamma}}, \boldsymbol{\gamma}, \in \mathbb{R}\setminus\{0\}$, 
there exists $\boldsymbol{\mu^*} \in \mathbb{R}$ and $\boldsymbol{\sigma^*} \in \mathbb{R}\setminus\{0\}$ such that $f(\bold{Z}; \boldsymbol{\mu^*},\boldsymbol{\sigma^*},\boldsymbol{\gamma},\boldsymbol{\beta}) = f(\bold{Z}; \boldsymbol{\tilde{\mu}},\boldsymbol{\tilde{\sigma}},\boldsymbol{\tilde{\gamma}},\boldsymbol{\tilde{\beta}})$:
By expanding the BN operation in Equation~\ref{eqn: appendix_bn}, we obtain:
\begin{align}
    f(\bold{Z}; \boldsymbol{\mu},\boldsymbol{\sigma},\boldsymbol{\gamma},\boldsymbol{\beta})
    &= \frac{\bold{Z}-\boldsymbol{\mu} + \boldsymbol{\beta\sigma} / \boldsymbol{\gamma}}{\boldsymbol{\sigma} / \boldsymbol{\gamma}} \label{eqn: appendix_bn_expand}
\end{align}
By equating the numerator (and denominator) in Equation~\ref{eqn: appendix_bn_expand} for the two expressions and rearranging terms, we can obtain $\boldsymbol{\sigma^*}$ and $\boldsymbol{\mu^*}$ to fulfill the desired equivalence:
\begin{align}
    \boldsymbol{\sigma^*} 
    &= \left( \frac{\boldsymbol{\gamma}}{\boldsymbol{\tilde{\gamma}}}\right) \boldsymbol{\tilde{\sigma}} 
    \hspace{0.5cm} \text{and} \hspace{0.5cm} 
    \boldsymbol{\mu^*}
    = \left( \frac{\boldsymbol{\tilde{\beta}}-\boldsymbol{\beta}}{\boldsymbol{\tilde{\gamma}}} \right) \boldsymbol{\tilde{\sigma}} + \boldsymbol{\tilde{\mu}} 
\end{align}

This implies that we can obtain the optimum of the function $f(\bold{Z}; \boldsymbol{\mu},\boldsymbol{\sigma},\boldsymbol{\gamma},\boldsymbol{\beta})$ with fixed $\mathbf{Z}$ by optimizing only over either $\{\boldsymbol{\gamma},\boldsymbol{\beta}\}$ or $\{\boldsymbol{\mu},\boldsymbol{\sigma}\}$. 
\section{Further Experiment Results}
\label{sec: further_experiment_results}

\subsection{Dataset Details}
\label{sec: dataset_details}
\textbf{PACS.}
PACS~\cite{li2018pacs} is an image classification dataset commonly used for domain generalization evaluation. It contains images from 4 styles or domains, namely Art painting, Cartoon, Photo and Sketch. Each domain has a total of 7 classes: dog, elephant, giraffe, guitar, horse, house and person.
We follow the setup in MixStyle~\cite{zhou2021mixstyle,mixstylerepo}, the method with state-of-the-art generalization performance on PACS, to produce the source ResNet-18 models for evaluation. We conduct leave-one-domain-out evaluation by treating each of the 4 domains as the target in turn, and treating the other 3 domains as source. Each experiment is run for 5 seeds following the original MixStyle evaluation~\cite{zhou2021mixstyle}.

\noindent\textbf{Camelyon17.}
Camelyon17 is a WILDS dataset~\cite{wilds} used to benchmark image classification under naturally-occurring domain shifts. The dataset comprises tissue patches from different hospitals for tumor detection i.e. binary classification. We directly use the ERM DenseNet-121 models trained over 10 seeds provided at \cite{wildsrepo} as our source models. It contains 2 classes in 2 domains.

\noindent\textbf{iWildCam.}
iWildCam is a WILDS dataset~\cite{wilds} for animal species classification using photos collected from camera traps at different locations.  We directly use the ERM ResNet-50 models trained over 3 seeds provided at \cite{wildsrepo} as our source models. The dataset contains a total of 182 classes in 2 domains.

\noindent\textbf{VisDA.}
VisDA~\cite{Peng2017VisDATV} is a popular image classification dataset to evaluate model performance under synthetic-to-real domain shift, and it contains 12 classes in 2 domains. 
We follow the setup in CSG~\cite{chen2021contrastive}, which achieved the state-of-the-art performance using synthetic rendering of 3D models as source domain and real images from Microsoft COCO as target domain. We directly use the CSG ResNet-101 model provided at \cite{csgrepo} as our source model, and run adaptation methods over 3 seeds.

\noindent\textbf{Office.}
Office~\cite{Saenko2010AdaptingVC} is a popular dataset for domain adaptation with 31 categories of office objects in 3 domains: Amazon (A), Webcam (W) and DSLR (D). We evaluate on all 6 domain pairs with ERM ResNet-50 source models trained with \cite{mixstylerepo} over 3 seeds.

\noindent\textbf{OfficeHome.}
OfficeHome~\cite{Venkateswara2017DeepHN} is a challenging dataset for domain adaptation with 65 categories of everyday objects in 4 domains: Art (Ar), Clipart (Cl), Product (Pr) and Real-World (Rw). We evaluate on all 12 domain pairs with ERM ResNet-50 source models trained with \cite{mixstylerepo} over 3 seeds.

\noindent\textbf{Segmentation datasets.} GTAV~\cite{richter2016gtav} is a benchmark synthetic dataset of driving-scene images from Grand Theft Auto V game. SYNTHIA~\cite{ros2016synthia} contains photo-realistic frames from a virtual city. For real-world datasets, Cityscapes~\cite{cordts2016cityscapes} contains images from 50 cities for semantic urban scene understanding, BDD-100K~\cite{yu2020bdd100k} is a driving dataset with diverse weather conditions and scene types, and Mapillary~\cite{neuhold2017mapillary} is a large-scale street-level dataset with images from around the world taken using different imaging devices under diverse conditions. We direct use the ResNet-50 models trained over 3 seeds from \cite{choi2021isw} as source models.

\subsection{Additional Results for Comparison with Test-Time Adaptation}
\label{sec: additional_results_test_time_adaptation}

In Table~\ref{tab:pacs_results} and \ref{tab:camelyon_iwildcam_visda_results}, we provide detailed results with standard deviations for comparison with test-time adaptation methods. The large variation in target performance in Camelyon17 is inherited from the original source models, and due to variability of the quality of the support set at $k=1$ which translates to only two support samples.
\begin{table}[ht]
\centering
\begin{adjustbox}{max width=\columnwidth}
\begin{tabular}{*{6}{l}}
\toprule[1pt]\midrule[0.3pt]
\textbf{Method} &  \multicolumn{1}{c}{\textbf{Art}}  & \multicolumn{1}{c}{\textbf{Cartoon}}   & \multicolumn{1}{c}{\textbf{Photo}} & \multicolumn{1}{c}{\textbf{Sketch}}  & \multicolumn{1}{c}{\textbf{Avg}} \\ \midrule
Source model (ERM)          & 76.4 \rpm 1.0     & 75.8 \rpm 1.0     & 96.0 \rpm 0.3     & 67.0 \rpm 1.3     & 78.8 \rpm 0.7 \\
+ Test-time BN              & 81.0 \rpm 0.4     & 79.8 \rpm 0.2     & 96.2 \rpm 0.4     & 67.5 \rpm 0.7     & 81.1 \rpm 0.2 \\
+ Tent (Adam)               & 83.5 \rpm 0.5     & 81.8 \rpm 0.6     & 96.8 \rpm 0.4     & 71.3 \rpm 1.0     & 83.4 \rpm 0.3 \\
+ Tent (SGD)                & 81.1 \rpm 0.4     & 79.6 \rpm 0.5     & 96.5 \rpm 0.5     & 68.2 \rpm 0.7     & 81.4 \rpm 0.2 \\ \midrule
+ LCCS (k=1)       & 77.9 \rpm 3.6     & 80.0 \rpm 1.0     & 95.9 \rpm 0.7     & 72.5 \rpm 4.3     & 81.6 \rpm 1.2 \\
+ LCCS (k=5)       & \underline{84.0 \rpm 1.5}     & \underline{85.0 \rpm 0.7}     & \underline{97.5 \rpm 0.2}     & \underline{76.7 \rpm 1.9}     & \underline{85.8 \rpm 0.4} \\
+ LCCS (k=10)      & \textbf{86.8 \rpm 1.2}     & \textbf{86.4 \rpm 0.5}     & \textbf{97.7 \rpm 0.3}     & \textbf{79.4 \rpm 1.2}     & \textbf{87.6 \rpm 0.3} \\
\\[-1.8ex]\hline 
\hline \\[-1.8ex] 
Source model (MixStyle)     & 83.9 \rpm 0.1     & 79.1 \rpm 0.9     & 95.8 \rpm 0.2     & 73.8 \rpm 2.0     & 83.1 \rpm 0.5 \\
+ Test-time BN              & 83.3 \rpm 0.3     & 82.1 \rpm 0.7     & 96.7 \rpm 0.2     & 74.9 \rpm 0.5     & 84.3 \rpm 0.2 \\
+ Tent (Adam)               & \underline{86.0 \rpm 0.7}     & 83.6 \rpm 0.4     & 96.8 \rpm 0.4     & 79.2 \rpm 0.6     & 86.4 \rpm 0.3 \\
+ Tent (SGD)                & 83.7 \rpm 0.6     & 82.0 \rpm 0.6     & 96.4 \rpm 0.4     & 75.6 \rpm 0.7     & 84.4 \rpm 0.3 \\ \midrule
+ LCCS (k=1)       & 82.2 \rpm 0.9     & 80.4 \rpm 0.9     & 95.9 \rpm 0.6     & 78.9 \rpm 3.1     & 84.4 \rpm 1.1 \\
+ LCCS (k=5)       & 85.7 \rpm 1.2     & \underline{85.5 \rpm 0.9}     & \underline{97.2 \rpm 0.2}     & \underline{80.0 \rpm 2.2}     & \underline{87.1 \rpm 0.5} \\
+ LCCS (k=10)      & \textbf{87.7 \rpm 1.0}     & \textbf{86.9 \rpm 0.6}     & \textbf{97.5 \rpm 0.3}     & \textbf{83.0 \rpm 1.1}     & \textbf{88.8 \rpm 0.6} \\
\midrule[0.3pt]\bottomrule[1pt]
\end{tabular}
\end{adjustbox}
\caption{Classification accuracy comparison with test-time adaptation on multi-domain PACS for 7-class classification. \label{tab:pacs_results}}
%\vspace{-3mm}
\end{table}

\begin{table}[ht]
\centering
\begin{adjustbox}{max width=0.7\linewidth}
\begin{tabular}{*{4}{l}}
\toprule[1pt]\midrule[0.3pt]
\textbf{Method}             & \multicolumn{1}{c}{\textbf{Camelyon17}}    & \multicolumn{1}{c}{\textbf{iWildCam}}  & \multicolumn{1}{c}{\textbf{VisDA}} \\ \midrule
Source model                & 70.3 \rpm 6.4                     & \underline{31.0 \rpm 1.3}     & 64.7 \\
+ Test-time BN              & 89.9 \rpm 1.5         & 30.5 \rpm 0.4                 & 60.7 \rpm 0.1 \\
+ Tent (Adam)               & 64.1 \rpm 12.2                    & 18.3 \rpm 0.4                 & 26.5 \rpm 2.2 \\
+ Tent (SGD)                & \textbf{91.4 \rpm 0.9}            & 29.9 \rpm 0.4                 & 65.7 \rpm 0.0 \\ \midrule
+ LCCS (k=1)       & 76.6 \rpm 12.4                    & \textbf{31.8 \rpm 1.0}        & 67.8 \rpm 1.1 \\
+ LCCS (k=5)       & 88.3 \rpm 4.0                     & -                             & \underline{76.0 \rpm 0.5} \\
+ LCCS (k=10)      & \underline{90.2 \rpm 1.6}         & -                             & \textbf{79.2 \rpm 0.4} \\
\midrule[0.3pt]\bottomrule[1pt]
\end{tabular}
\end{adjustbox}
\caption{Classification performance comparison with test-time adaptation for binary classification on Camelyon17, 182-class classification on iWildCam, and 12-class classification on VisDA. \label{tab:camelyon_iwildcam_visda_results}}
\end{table}

We show two scenarios where test-time adaptation produces unreliable predictions on the multi-domain PACS dataset in Table~\ref{tab:pacs_failure_cases}. These two scenarios correspond to the challenges discussed in Section~\ref{sec: challenges_test_time_bn}, namely requirement for large mini-batch size and even class distribution in each mini-batch. When test mini-batch size is small at 8, both Test-time BN and Tent performs worse than the original source model. When test samples are sequentially ordered by class instead of randomly shuffled, performance severely degrades to approximately $37\%$ accuracy. Smaller extent of class imbalance when $\alpha=10$ and $\alpha=100$, where $\alpha$ is the sample size ratio of the largest to the smallest class, also reduces effectiveness of test-time adaptation. In contrast, all model parameters are frozen after our few-shot adaptation setting, hence the model adapted by our proposed method is not affected by mini-batch size and class-distribution at testing time.
\begin{table}[ht]
\centering

\subfigure[Effect of mini-batch size]{
\begin{adjustbox}{max width=\columnwidth}
\begin{tabular}{lc*{5}{l}}
\toprule[1pt]\midrule[0.3pt]
\textbf{Method} & \textbf{Test batch}   & \multicolumn{1}{c}{\textbf{Art}}  & \multicolumn{1}{c}{\textbf{Cartoon}}   & \multicolumn{1}{c}{\textbf{Photo}} & \multicolumn{1}{c}{\textbf{Sketch}}  & \multicolumn{1}{c}{\textbf{Avg}} \\ \midrule
\multicolumn{7}{l}{\underline{Random shuffling of test samples}} \\ 
Source model  & any   & 83.9 \rpm 0.1     & 79.1 \rpm 0.9     & 95.8 \rpm 0.2     & 73.8 \rpm 2.0     & 83.1 \rpm 0.5 \\
+ Test-time BN          & 8     & 78.4 \rpm 0.4     & 76.6 \rpm 0.5     & 89.5 \rpm 0.3     & 70.0 \rpm 0.5     & 78.6 \rpm 0.2 \\
+ Test-time BN          & 32    & 82.5 \rpm 0.5     & 81.0 \rpm 0.7     & 95.3 \rpm 0.2     & 74.4 \rpm 0.7     & 83.3 \rpm 0.1 \\
+ Test-time BN          & 64    & 83.0 \rpm 0.4     & 82.0 \rpm 0.6     & 96.5 \rpm 0.3     & 74.8 \rpm 0.6     & 84.1 \rpm 0.3 \\
+ Tent (Adam)           & 8     & 81.1 \rpm 0.7     & 79.3 \rpm 1.1     & 91.4 \rpm 0.8     & 72.7 \rpm 1.7     & 81.1 \rpm 0.6 \\
+ Tent (Adam)           & 32    & \underline{85.8 \rpm 0.5}     & 84.0 \rpm 0.5     & 96.5 \rpm 0.2     & 78.2 \rpm 0.7     & 86.1 \rpm 0.3 \\
+ Tent (Adam)           & 64    & \underline{85.8 \rpm 0.6}     & 83.1 \rpm 0.8     & 96.6 \rpm 0.4     & \underline{80.1 \rpm 0.6}     & 86.4 \rpm 0.4 \\ \midrule
+ LCCS ($k=1$) & any   & 82.2 \rpm 0.9     & 80.4 \rpm 0.9     & 95.9 \rpm 0.6     & 78.9 \rpm 3.1     & 84.4 \rpm 1.1 \\
+ LCCS ($k=5$) & any   & 85.7 \rpm 1.2     & \underline{85.5 \rpm 0.9}     & \underline{97.2 \rpm 0.2}     & 80.0 \rpm 2.2     & \underline{87.1 \rpm 0.5} \\
+ LCCS ($k=10$)& any   & \textbf{87.7 \rpm 1.0}     & \textbf{86.9 \rpm 0.6}     & \textbf{97.5 \rpm 0.3}     & \textbf{83.0 \rpm 1.1}     & \textbf{88.8 \rpm 0.6} \\
\midrule[0.3pt]\bottomrule[1pt]
\end{tabular}
\end{adjustbox}
}

\subfigure[Effect of imbalanced class distribution]{
\begin{adjustbox}{max width=\columnwidth}
\begin{tabular}{lc*{5}{l}}
\toprule[1pt]\midrule[0.3pt]
\textbf{Method} & \textbf{Test batch}   & \multicolumn{1}{c}{\textbf{Art}}  & \multicolumn{1}{c}{\textbf{Cartoon}}   & \multicolumn{1}{c}{\textbf{Photo}} & \multicolumn{1}{c}{\textbf{Sketch}}  & \multicolumn{1}{c}{\textbf{Avg}} \\ \midrule
\multicolumn{7}{l}{\underline{Sequential ordering of test samples by class}} \\ 
Source model  & any   & 83.9 \rpm 0.1     & 79.1 \rpm 0.9     & 95.8 \rpm 0.2     & 73.8 \rpm 2.0     & 83.1 \rpm 0.5 \\
+ Test-time BN          & 128   & 38.4 \rpm 0.2     & 38.7 \rpm 0.6     & 45.8 \rpm 0.5     & 27.4 \rpm 0.4     & 37.6 \rpm 0.3 \\
+ Tent (Adam)           & 128   & 37.9 \rpm 0.3     & 38.5 \rpm 0.6     & 45.8 \rpm 0.4     & 26.2 \rpm 0.3     & 37.1 \rpm 0.3 \\ \midrule
+ LCCS ($k=1$) & any   & 82.2 \rpm 0.9     & 80.4 \rpm 0.9     & 95.9 \rpm 0.6     & 78.9 \rpm 3.1     & 84.4 \rpm 1.1 \\
+ LCCS ($k=5$) & any   & \underline{85.7 \rpm 1.2}     & \underline{85.5 \rpm 0.9}     & \underline{97.2 \rpm 0.2}     & \underline{80.0 \rpm 2.2}     & \underline{87.1 \rpm 0.5} \\
+ LCCS ($k=10$)& any   & \textbf{87.7 \rpm 1.0}     & \textbf{86.9 \rpm 0.6}     & \textbf{97.5 \rpm 0.3}     & \textbf{83.0 \rpm 1.1}     & \textbf{88.8 \rpm 0.6} \\
\\[-1.8ex]\hline 
\hline \\[-1.8ex] 

\multicolumn{7}{l}{\underline{Long-tailed class distribution, $\alpha=10$}} \\ 
Source model  & any   & 79.5 \rpm 1.2     & 78.2 \rpm 1.3     & 93.2 \rpm 1.0     & 68.8 \rpm 2.2     & 79.9 \rpm 0.6 \\
+ Test-time BN          & 128   & 80.9 \rpm 0.6     & 76.1 \rpm 0.3     & 89.6 \rpm 0.8     & 67.8 \rpm 0.6     & 78.6 \rpm 0.4 \\
+ Tent (Adam)           & 128   & 82.2 \rpm 0.8     & 76.8 \rpm 0.6     & 90.0 \rpm 0.8     & 71.1 \rpm 0.8     & 80.0 \rpm 0.3\\ \midrule
+ LCCS ($k=1$) & any   & 80.8 \rpm 2.9     & 77.2 \rpm 4.3     & 93.1 \rpm 0.9     & 74.1 \rpm 2.8     & 81.3 \rpm 1.2 \\
+ LCCS ($k=5$) & any   & \underline{85.1 \rpm 2.1}     & \underline{83.1 \rpm 3.0}     & \underline{94.2 \rpm 2.1}     & \underline{77.2 \rpm 2.2}     & \underline{84.9 \rpm 0.7} \\
+ LCCS ($k=10$)& any   & \textbf{85.7 \rpm 1.8}     & \textbf{85.3 \rpm 1.9}     & \textbf{96.0 \rpm 0.8}     & \textbf{80.3 \rpm 0.5}     & \textbf{86.8 \rpm 0.8} \\
\\[-1.8ex]\hline 
\hline \\[-1.8ex] 

\multicolumn{7}{l}{\underline{Long-tailed class distribution, $\alpha=100$}} \\
Source model  & any   & 77.7 \rpm 1.9     & 74.9 \rpm 2.0     & 92.7 \rpm 1.1     & 62.2 \rpm 2.8     & 76.9 \rpm 0.8 \\
+ Test-time BN          & 128   & 70.3 \rpm 0.7     & 62.6 \rpm 0.2     & 79.5 \rpm 1.1     & 54.0 \rpm 1.1     & 66.6 \rpm 0.6 \\
+ Tent (Adam)           & 128   & 71.8 \rpm 0.6     & 63.2 \rpm 0.4     & 80.0 \rpm 1.2     & 55.7 \rpm 1.4     & 67.7 \rpm 0.7 \\ \midrule
+ LCCS ($k=1$) & any   & 79.7 \rpm 3.2     & 73.0 \rpm 7.2     & \underline{93.4 \rpm 1.5}     & 68.5 \rpm 5.0     & 78.4 \rpm 1.9 \\
+ LCCS ($k=5$) & any   & \underline{83.4 \rpm 3.3}     & \underline{80.1 \rpm 4.5}     & 92.5 \rpm 3.4     & \underline{71.7 \rpm 3.3}     & \underline{81.9 \rpm 1.2} \\
+ LCCS ($k=10$)& any   & \textbf{83.8 \rpm 3.1}     & \textbf{82.7 \rpm 2.3}     & \textbf{95.5 \rpm 1.1}     & \textbf{75.1 \rpm 1.4}     & \textbf{84.3 \rpm 0.9} \\
\midrule[0.3pt]\bottomrule[1pt]
\end{tabular}
\end{adjustbox}}
\caption{Classification accuracy on PACS for streaming settings where test-time adaptation can result in worse performance than the original source model. \label{tab:pacs_failure_cases}}
\end{table}

\subsection{Additional Results for Comparison with Few-Shot Transfer Learning}
\label{sec: additional_results_source_free_adaptation}

In Table~\ref{tab:full_results_source_free_few_shot}, we provide detailed results with standard deviations for comparison with source-free few-shot transfer learning methods. The large variation in target performance in Camelyon17 is inherited from the original source models, and due to variability of the quality of the support set at $k=1$ which translates to only two support samples. We observe that standard deviations tend to be larger at $k=1$ for all datasets due to variability of the quality of the extremely small support set, and this is especially so for FLUTE and finetuning LCCS with NCC classifier because the centroid classifier depends directly on how representative the support set is of the entire target domain dataset.
\begin{table*}[ht]
\centering
\begin{adjustbox}{max width=0.8\textwidth}
\setlength\tabcolsep{4pt} % default value: 6pt
\begin{tabular}{*{10}{l}}
\toprule[1pt]\midrule[0.3pt]
\textbf{Method}             & \multicolumn{3}{c}{\textbf{PACS-Art}}  & \multicolumn{3}{c}{\textbf{PACS-Cartoon}}  & \multicolumn{3}{c}{\textbf{PACS-Photo}} \\ \cmidrule{2-10}
\multicolumn{1}{r}{$k=$}    & \multicolumn{1}{c}{1} & \multicolumn{1}{c}{5} & \multicolumn{1}{c}{10}
                            & \multicolumn{1}{c}{1} & \multicolumn{1}{c}{5} & \multicolumn{1}{c}{10}
                            & \multicolumn{1}{c}{1} & \multicolumn{1}{c}{5} & \multicolumn{1}{c}{10} \\ \cmidrule(lr){2-4} \cmidrule(lr){5-7} \cmidrule(lr){8-10} 
AdaBN                       & 80.7 \rpm 2.0 & 84.5 \rpm 0.7 & 85.0 \rpm 0.6            & 79.9 \rpm 0.9 & 83.2 \rpm 0.8 & 83.5 \rpm 0.8                
                            & 95.2 \rpm 0.6 & 96.0 \rpm 0.8 & 96.0 \rpm 0.7\\
finetune BN                 & 74.2 \rpm 3.6 & 80.9 \rpm 0.9 & 83.2 \rpm 1.4            & 77.9 \rpm 1.8 & 82.0 \rpm 1.8 & 83.1 \rpm 1.4                
                            & 92.4 \rpm 0.6 & 95.9 \rpm 0.7 & 96.2 \rpm 0.5 \\
finetune classifier         & 81.7 \rpm 0.7 & 83.7 \rpm 0.6 & 84.2 \rpm 0.3            & 79.2 \rpm 0.6 & 80.5 \rpm 0.6 & 80.5 \rpm 0.6                
                            & 95.6 \rpm 0.6 & 96.0 \rpm 0.3 & 96.1 \rpm 0.3 \\
finetune feat. extractor    & \textbf{83.3 \rpm 0.8} & 86.1 \rpm 1.3 & 86.1 \rpm 1.0            & \textbf{81.8 \rpm 1.0} & 84.0 \rpm 1.5 & 85.4 \rpm 0.8                
                            & \underline{95.9 \rpm 0.6} & 96.4 \rpm 0.5 & 96.6 \rpm 0.7 \\
$L^2$                       & \textbf{83.3 \rpm 0.9} & \textbf{85.8 \rpm 0.9} & 85.6 \rpm 0.8            & \textbf{81.8 \rpm 0.7} & 83.4 \rpm 0.4 & 84.1 \rpm 0.9                
                            & \textbf{96.1 \rpm 0.4} & 96.6 \rpm 0.5 & 96.4 \rpm 0.6 \\
$L^2$-SP                    & \textbf{83.3 \rpm 0.9} & \textbf{85.8 \rpm 0.9} & 85.6 \rpm 0.8            & \textbf{81.8 \rpm 0.7} & 83.4 \rpm 0.4 & 84.1 \rpm 0.9               
                            & \textbf{96.1 \rpm 0.4} & 96.6 \rpm 0.5 & 96.4 \rpm 0.6 \\ 
DELTA                       & \textbf{83.3 \rpm 1.0} & \textbf{85.8 \rpm 0.8} & 85.6 \rpm 0.7            & \textbf{81.8 \rpm 0.7} & 83.3 \rpm 0.3 & 83.8 \rpm 0.9               
                            & \textbf{96.1 \rpm 0.4} & \underline{96.7 \rpm 0.5} & 96.5 \rpm 0.7 \\
Late Fusion                 & \underline{83.0 \rpm 0.6} & 83.8 \rpm 0.7 & 83.8 \rpm 0.5            & 79.8 \rpm 0.2 & 79.9 \rpm 0.5 & 79.7 \rpm 0.6                
                            & 95.8 \rpm 0.4 & 95.9 \rpm 0.3 & 96.0 \rpm 0.2 \\
FLUTE                       & 67.0 \rpm 9.7 & 83.6 \rpm 3.0 & \underline{87.2 \rpm 0.8}            & 73.5 \rpm 6.4 & \underline{84.7 \rpm 0.7} & \underline{86.1 \rpm 0.5}                
                            & 90.3 \rpm 4.1 & 96.3 \rpm 0.9 & \underline{97.2 \rpm 0.2} \\ \midrule
LCCS                        & 82.2 \rpm 0.9 & \underline{85.7 \rpm 1.2} & \textbf{87.7 \rpm 1.0}            & \underline{80.4 \rpm 0.9} & \textbf{85.5 \rpm 0.9} & \textbf{86.9 \rpm 0.6}               
                            & \underline{95.9 \rpm 0.6} & \textbf{97.2 \rpm 0.2} & \textbf{97.5 \rpm 0.3} \\
\midrule[0.3pt]\bottomrule[1pt]
\end{tabular}
\end{adjustbox}

\medskip

\begin{adjustbox}{max width=0.8\textwidth}
\setlength\tabcolsep{4pt} % default value: 6pt
\begin{tabular}{*{10}{l}}
\toprule[1pt]\midrule[0.3pt]
\textbf{Method}             & \multicolumn{3}{c}{\textbf{PACS-Sketch}} & \multicolumn{3}{c}{\textbf{Camelyon17}}  & \multicolumn{3}{c}{\textbf{VisDA}} \\ \cmidrule{2-10}
\multicolumn{1}{r}{$k=$}    & \multicolumn{1}{c}{1} & \multicolumn{1}{c}{5} & \multicolumn{1}{c}{10}
                            & \multicolumn{1}{c}{1} & \multicolumn{1}{c}{5} & \multicolumn{1}{c}{10}
                            & \multicolumn{1}{c}{1} & \multicolumn{1}{c}{5} & \multicolumn{1}{c}{10} \\ \cmidrule(lr){2-4} \cmidrule(lr){5-7} \cmidrule(lr){8-10}
AdaBN                       & 75.8 \rpm 2.8 & 78.5 \rpm 0.6 & 78.7 \rpm 0.6            & 72.9 \rpm 14.3 & 87.8 \rpm 5.4 & \underline{90.2 \rpm 1.4} 
                            & 56.5 \rpm 0.7 & 60.9 \rpm 0.7 & 61.8 \rpm 0.3 \\
finetune BN                 & 71.7 \rpm 3.1 & 78.6 \rpm 2.5 & 79.0 \rpm 2.2            & 72.6 \rpm 7.2 & 87.7 \rpm 6.2 & 90.1 \rpm 3.5              
                            & 59.1 \rpm 0.6 & \underline{70.9 \rpm 3.0} & 74.9 \rpm 1.5 \\
finetune classifier         & 73.2 \rpm 1.1 & 74.6 \rpm 1.0 & 74.6 \rpm 1.4            & 70.5 \rpm 6.1 & 70.4 \rpm 6.1 & 70.5 \rpm 6.0              
                            & \underline{67.6 \rpm 1.4} & 69.7 \rpm 0.9 & \underline{77.4 \rpm 0.6} \\
finetune feat. extractor    & 73.4 \rpm 5.5 & 77.5 \rpm 2.5 & 76.3 \rpm 1.6            & \underline{79.3 \rpm 11.0} & 86.5 \rpm 5.4 & 88.3 \rpm 4.4 
                            & 67.3 \rpm 0.7 & 68.4 \rpm 0.4 & 74.7 \rpm 0.7 \\
$L^2$                       & 76.5 \rpm 2.2 & 77.5 \rpm 2.2 & 76.3 \rpm 1.1            & \textbf{79.6 \rpm 8.3} & \underline{88.2 \rpm 3.8} & 89.5 \rpm 2.8 
                            & 66.0 \rpm 0.6 & 66.4 \rpm 0.5 & 69.6 \rpm 0.5 \\
$L^2$-SP                    & 76.5 \rpm 2.2 & 77.5 \rpm 2.2 & 76.3 \rpm 1.1            & \textbf{79.6 \rpm 8.3} & \underline{88.2 \rpm 3.8} & 89.5 \rpm 2.8 
                            & 66.0 \rpm 0.6 & 66.4 \rpm 0.5 & 69.6 \rpm 0.5 \\ 
DELTA                       & \underline{76.6 \rpm 2.3} & 77.4 \rpm 2.1 & 75.4 \rpm 1.3            & \textbf{79.6 \rpm 8.3} & \underline{88.2 \rpm 3.8} & 89.5 \rpm 2.8 
                            & 65.9 \rpm 0.7 & 66.5 \rpm 0.5 & 70.1 \rpm 0.6 \\
Late Fusion                 & 74.3 \rpm 1.7 & 75.1 \rpm 1.3 & 74.9 \rpm 1.1            & 70.4 \rpm 6.1 & 70.4 \rpm 6.1 & 70.5 \rpm 6.0                
                            & 67.2 \rpm 1.2 & 69.8 \rpm 1.1 & 74.5 \rpm 0.6 \\
FLUTE                       & 62.8 \rpm 9.3 & \underline{78.7 \rpm 3.4} & \underline{81.7 \rpm 1.5}            & 73.1 \rpm 7.5 & 86.5 \rpm 7.0 & \textbf{90.9 \rpm 3.3}               
                            & 48.3 \rpm 1.9 & 67.1 \rpm 1.9 & 65.7 \rpm 1.9 \\ \midrule
LCCS                        & \textbf{78.9 \rpm 3.1} & \textbf{80.0 \rpm 2.2} & \textbf{83.0 \rpm 1.1}            & 76.6 \rpm 12.4 & \textbf{88.3 \rpm 4.0} & \underline{90.2 \rpm 1.6}               
                            & \textbf{67.8 \rpm 1.1}  & \textbf{76.0 \rpm 0.5} & \textbf{79.2 \rpm 0.4} \\
\midrule[0.3pt]\bottomrule[1pt]
\end{tabular}
\end{adjustbox}

\caption{Classification performance comparison with few-shot transfer learning. \label{tab:full_results_source_free_few_shot}}
%\vspace{-3mm}
\end{table*}

\subsection{Additional Results for Comparison with Source-Free Unsupervised Domain Adaptation}
\label{sec: additional_results_other_DA}

Additionally, we compare with state-of-the-art source-free UDA method SHOT~\cite{liang2020shot}, which adapts on the entire unlabeled target dataset, on OfficeHome. From Table~\ref{tab:officehome_results_resnet50}, SHOT performs better than finetuning LCCS in most domain pairs. This reflects that adapting with limited samples is a challenging task, and we will explore strategies to further close the performance gap in future work. We note though that finetuning LCCS with finetuned linear classifier (source classifier finetuned for 200 epochs) performs better than the other few-shot adaptation methods evaluated. 
\begin{table*}[ht]
\centering
\begin{adjustbox}{max width=\textwidth}
\begin{tabular}{l*{14}{c}}
\toprule[1pt]\midrule[0.3pt]
\textbf{Method}         & \multicolumn{1}{c}{$\boldsymbol{k}$} 
                                                & $\mathbf{A\rightarrow C}$ & $\mathbf{A\rightarrow P}$ & $\mathbf{A\rightarrow R}$  
                                                & $\mathbf{C\rightarrow A}$ & $\mathbf{C\rightarrow P}$    & $\mathbf{C\rightarrow R}$ 
                                                & $\mathbf{P\rightarrow A}$ & $\mathbf{P\rightarrow C}$    & $\mathbf{P\rightarrow R}$
                                                & $\mathbf{R\rightarrow A}$ & $\mathbf{R\rightarrow C}$    & $\mathbf{R\rightarrow P}$ & \textbf{Avg}\\ \midrule
SHOT                    & all$^\dagger$         & \underline{57.1} & \textbf{78.1} & \textbf{81.5} & \textbf{68.0} & \textbf{78.2} & \textbf{78.1} & \textbf{67.4} & \textbf{54.9} & \textbf{82.2} & \textbf{73.3} & \underline{58.8} & \textbf{84.3} & \textbf{71.8}\\
SFDA                    & all$^\dagger$         & 48.4 & 73.4 & 76.9 & \underline{64.3} & 69.8 & \underline{71.7} & \underline{62.7} & 45.3 & \underline{76.6} & \underline{69.8} & 50.5 & 79.0 & 65.7 \\ 
AdaBN                   & all$^\dagger$         & 50.9 & 63.1 & 72.3 & 53.2 & 62.0 & 63.4 & 52.2 & 49.8 & 71.5 & 66.1 & 56.1 & 77.1 & 61.5 \\
L$^2$                   & 5                     & 52.5 & 66.1 & 73.4 & 56.1 & 64.9 & 65.2 & 54.7 & 50.0 & 73.4 & 67.8 & 57.0 & 78.9 & 63.3 \\
FLUTE                   & 5                     & 49.0 & 70.1 & 68.2 & 53.8 & 69.3 & 65.1 & 53.2 & 46.8 & 70.8 & 59.4 & 51.7 & 77.3 & 61.2 \\ \midrule
LCCS$^*$      & 5                     & \textbf{57.6} & \underline{74.5} & \underline{77.0} & 60.0 & \underline{71.5} & 70.9 & 59.2 & \underline{54.7} & 75.9 & 69.2 & \textbf{61.2} & \underline{81.5} & \underline{67.8} \\
\midrule[0.3pt]\bottomrule[1pt]
\end{tabular}
\end{adjustbox}
\caption{Classification accuracy for 65-class classification on OfficeHome. $\dagger$ denotes target samples are unlabeled. $*$ denotes linear classifier is finetuned on support set. \label{tab:officehome_results_resnet50}}
\end{table*}

\subsection{Additional Results for Semantic Segmentation}
\label{sec: additional_segmentation}

\begin{figure*}[htb]
    \centering

    \includegraphics[width=0.24\linewidth]{figures/segmentation/frankfurt_000000_001016_leftImg8bit.png}
    \hfill
    \includegraphics[width=0.24\linewidth]{figures/segmentation/frankfurt_000000_001016_gtFine_color.png}
    \hfill
    \includegraphics[width=0.24\linewidth]{figures/segmentation/frankfurt_000000_001016_leftImg8bit_color_isw.png}
    \hfill
    \includegraphics[width=0.24\linewidth]{figures/segmentation/frankfurt_000000_001016_leftImg8bit_color_lccs.png}    
    \hfill
    
    \includegraphics[width=0.24\linewidth]{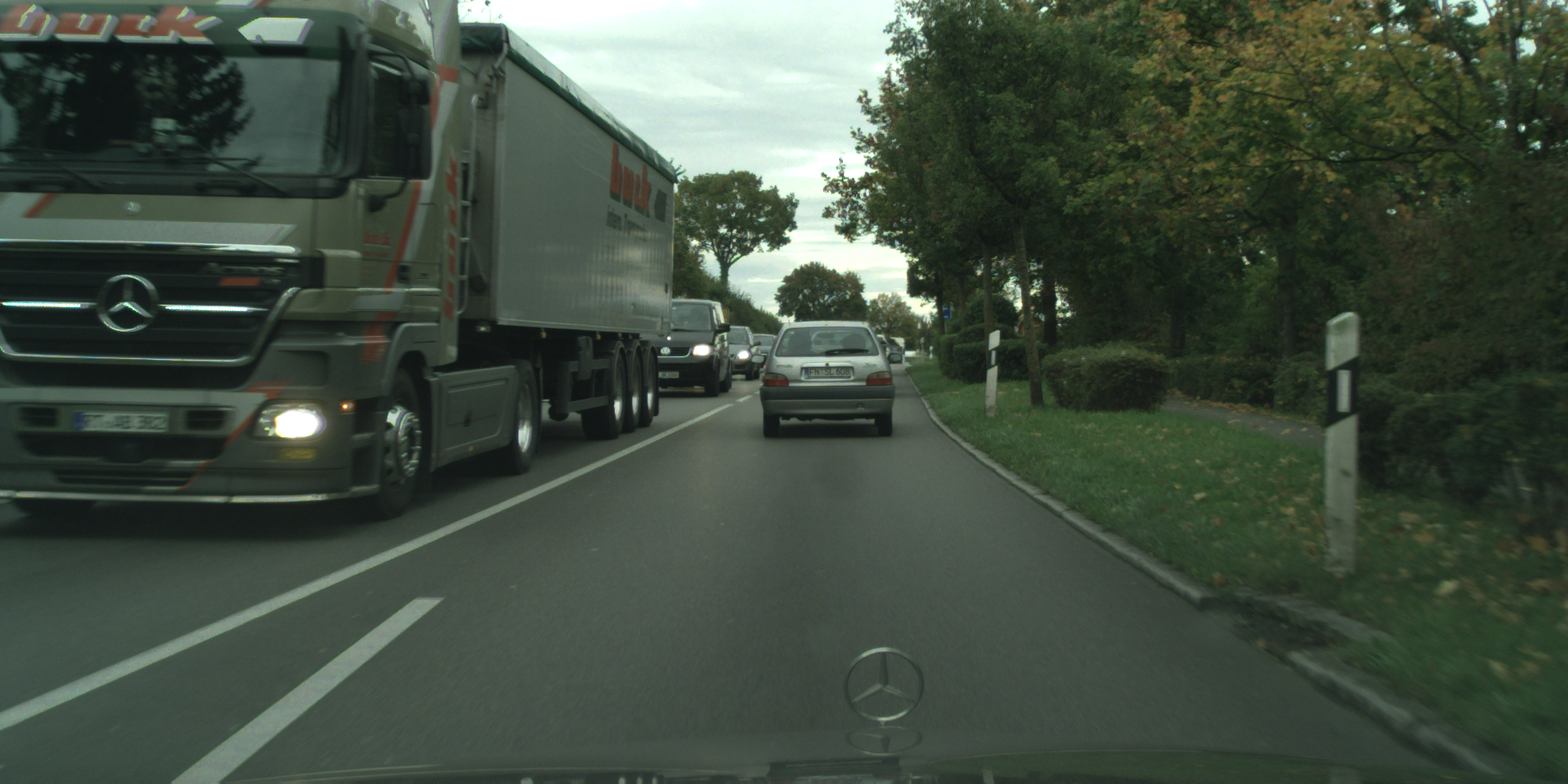}
    \hfill
    \includegraphics[width=0.24\linewidth]{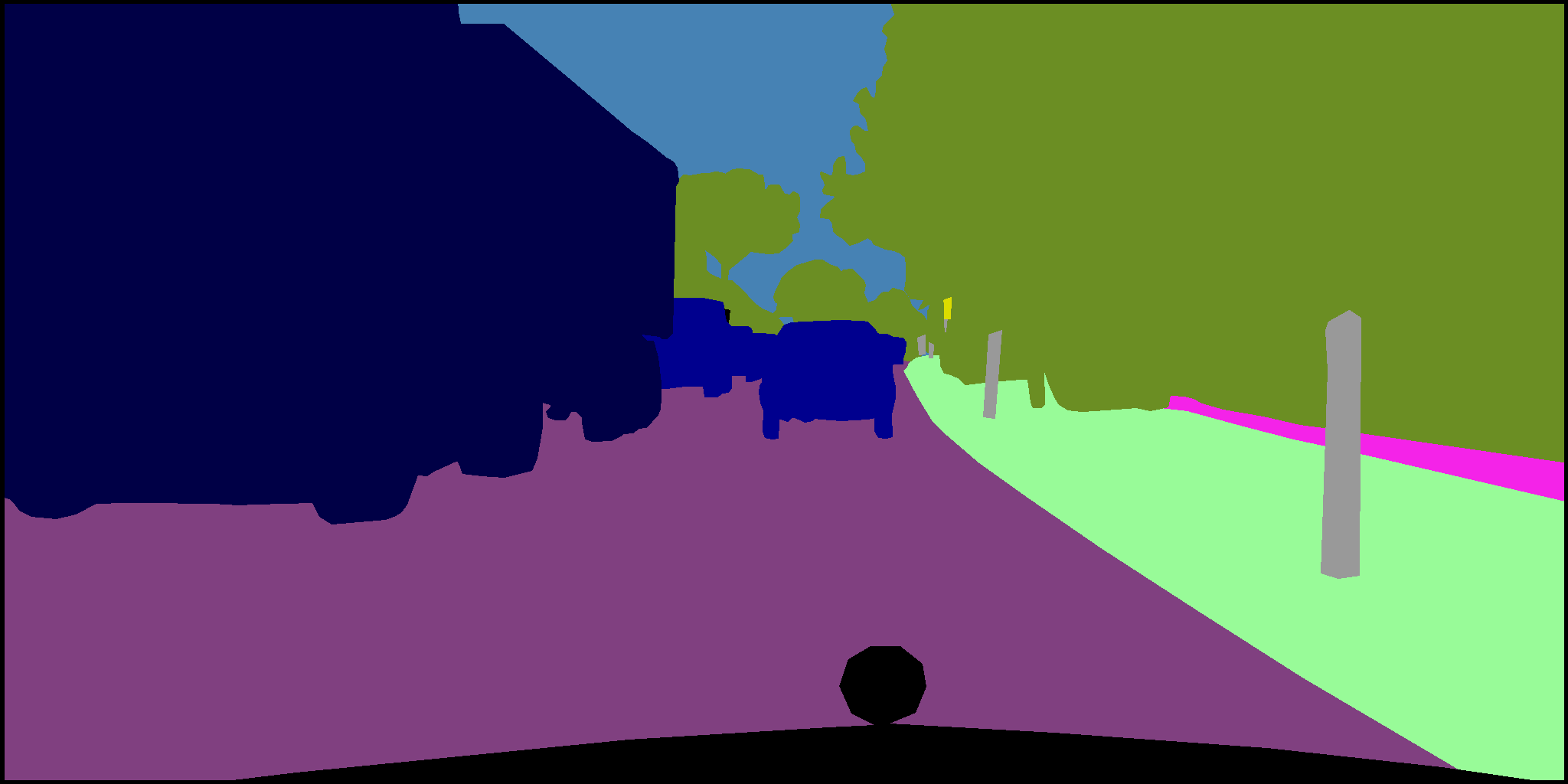}
    \hfill
    \includegraphics[width=0.24\linewidth]{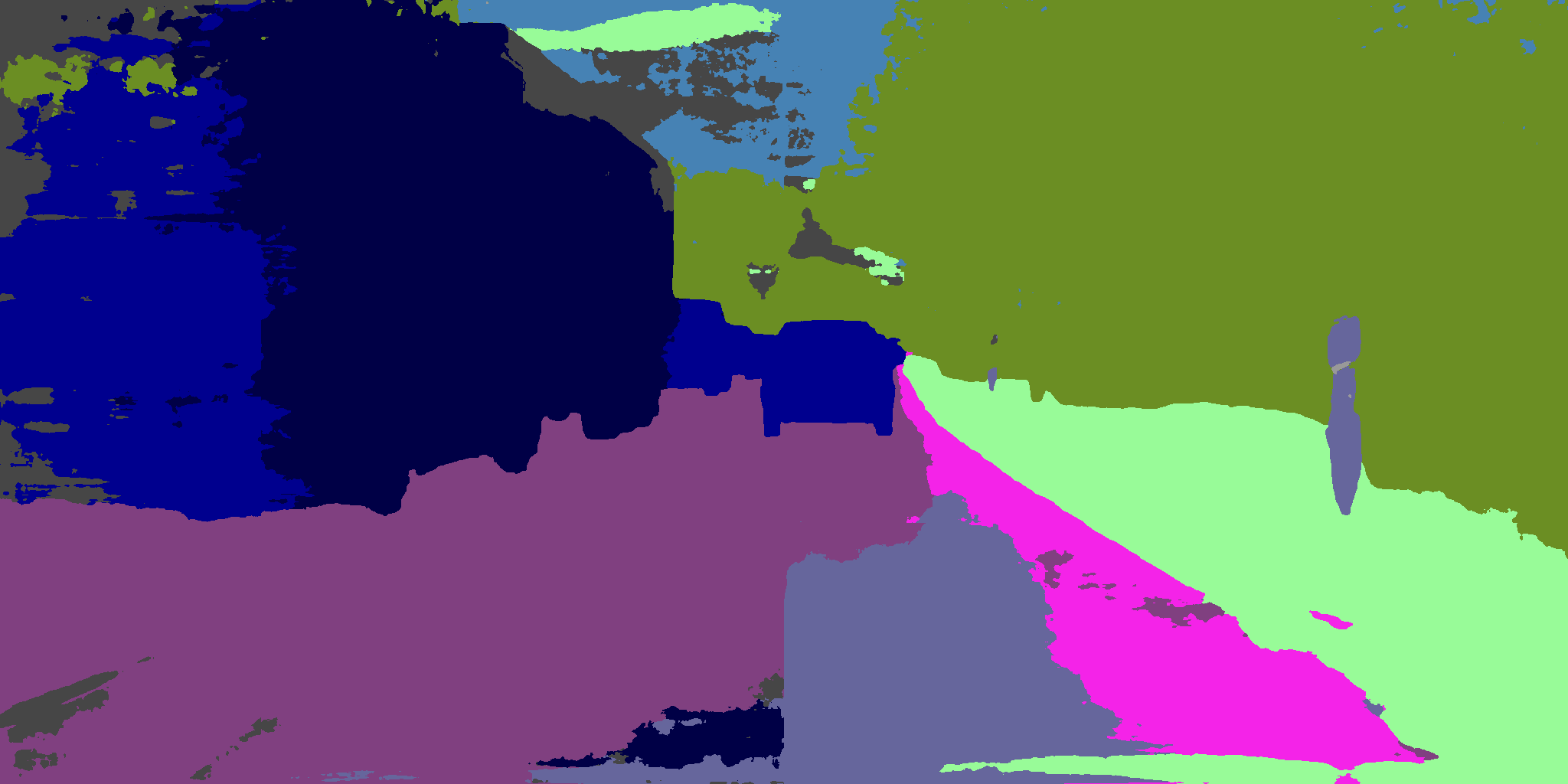}
    \hfill
    \includegraphics[width=0.24\linewidth]{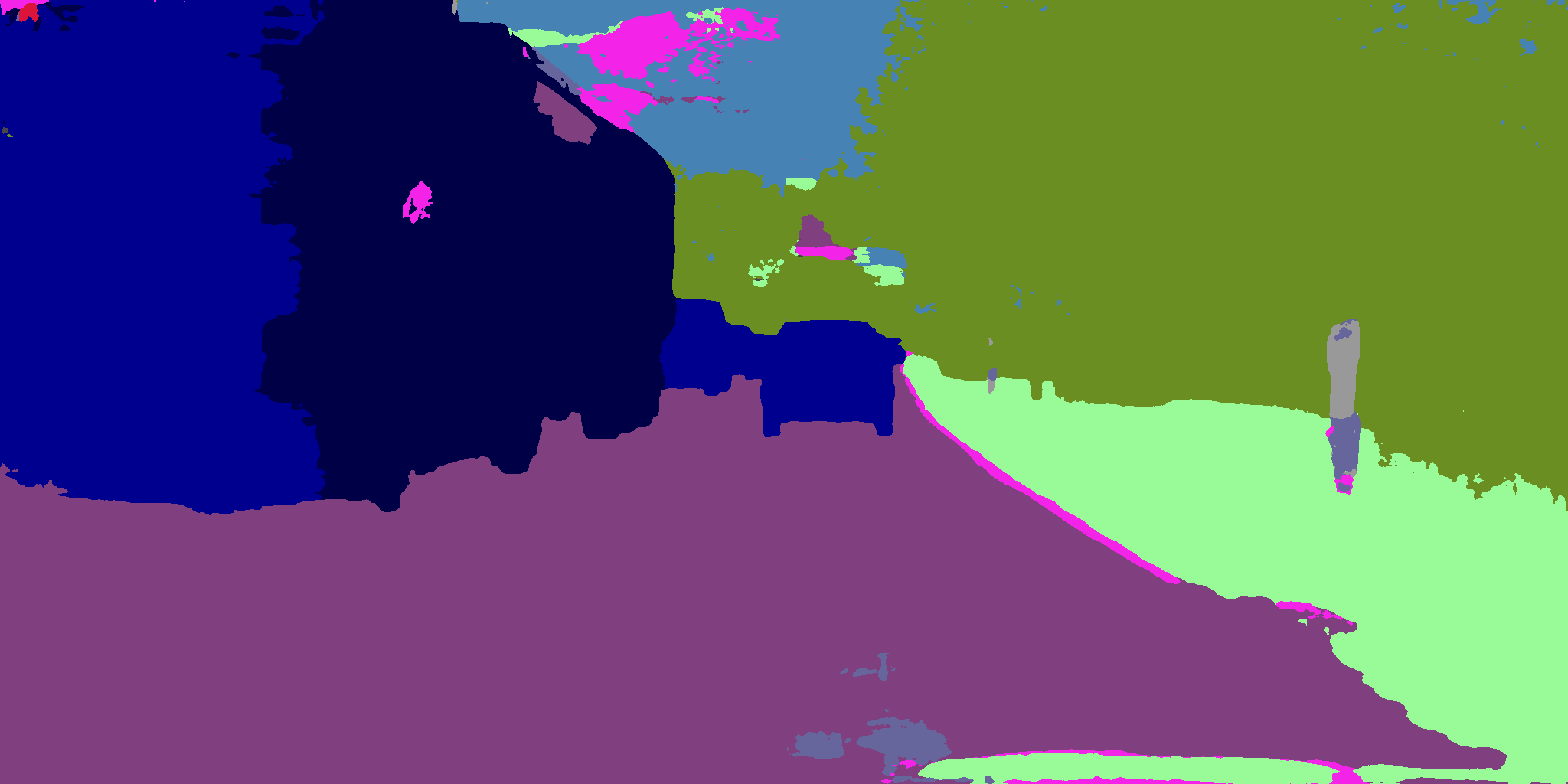}    
    \hfill    
    
    \subfigure[Target image]{\includegraphics[width=0.24\linewidth]{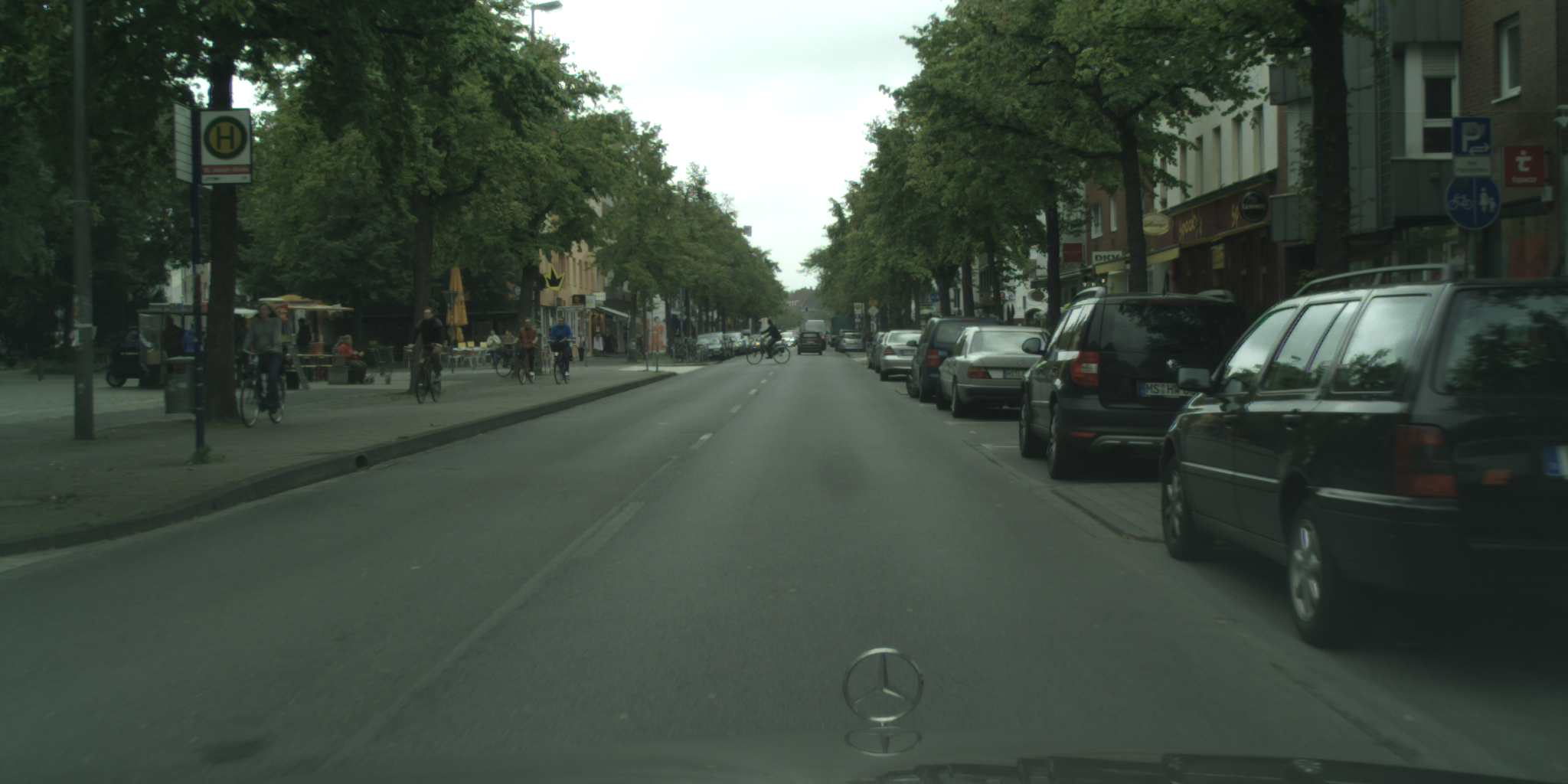}}
    \hfill
    \subfigure[Ground truth]{\includegraphics[width=0.24\linewidth]{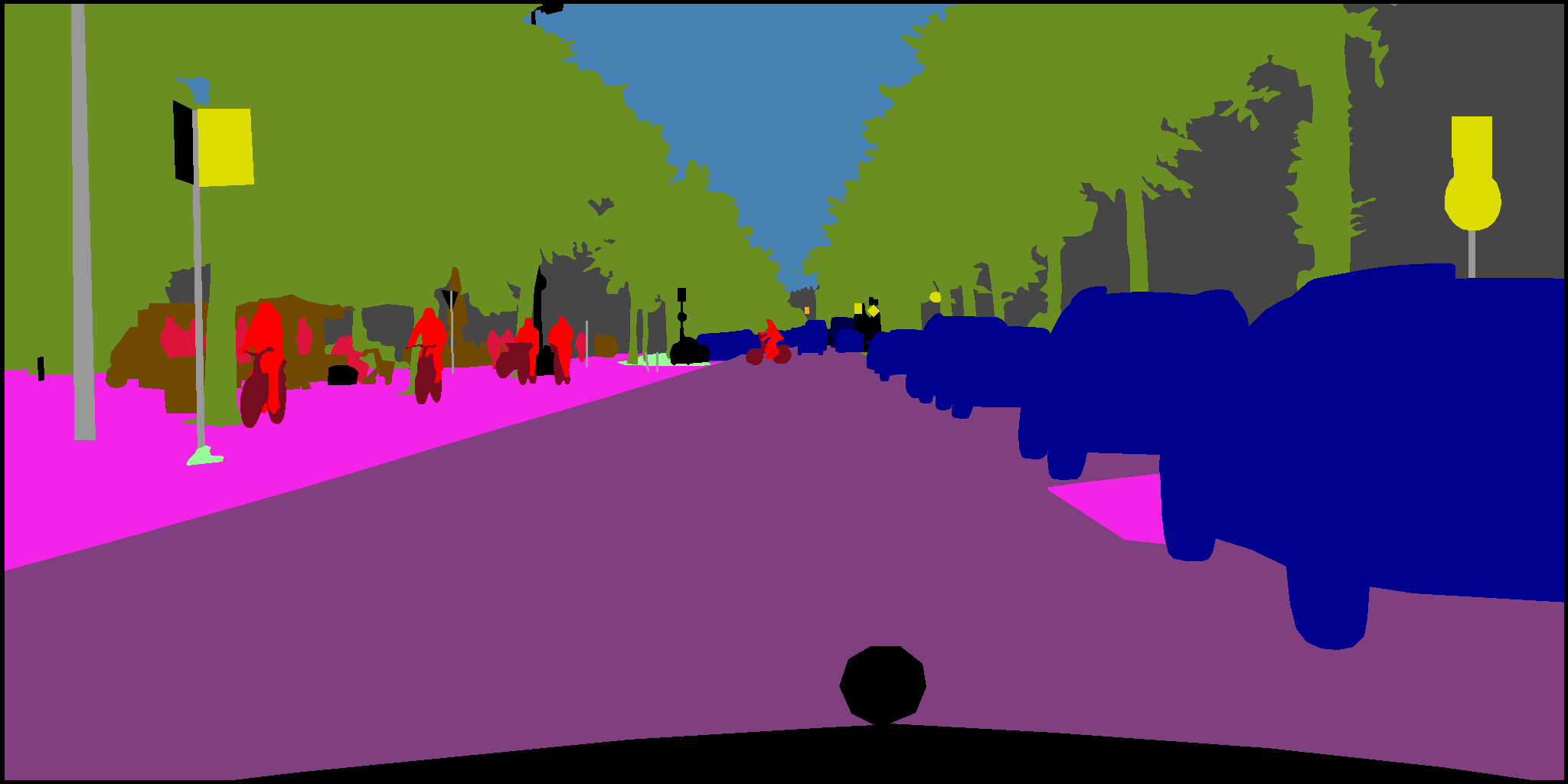}}
    \hfill
    \subfigure[ISW (SOTA)]{\includegraphics[width=0.24\linewidth]{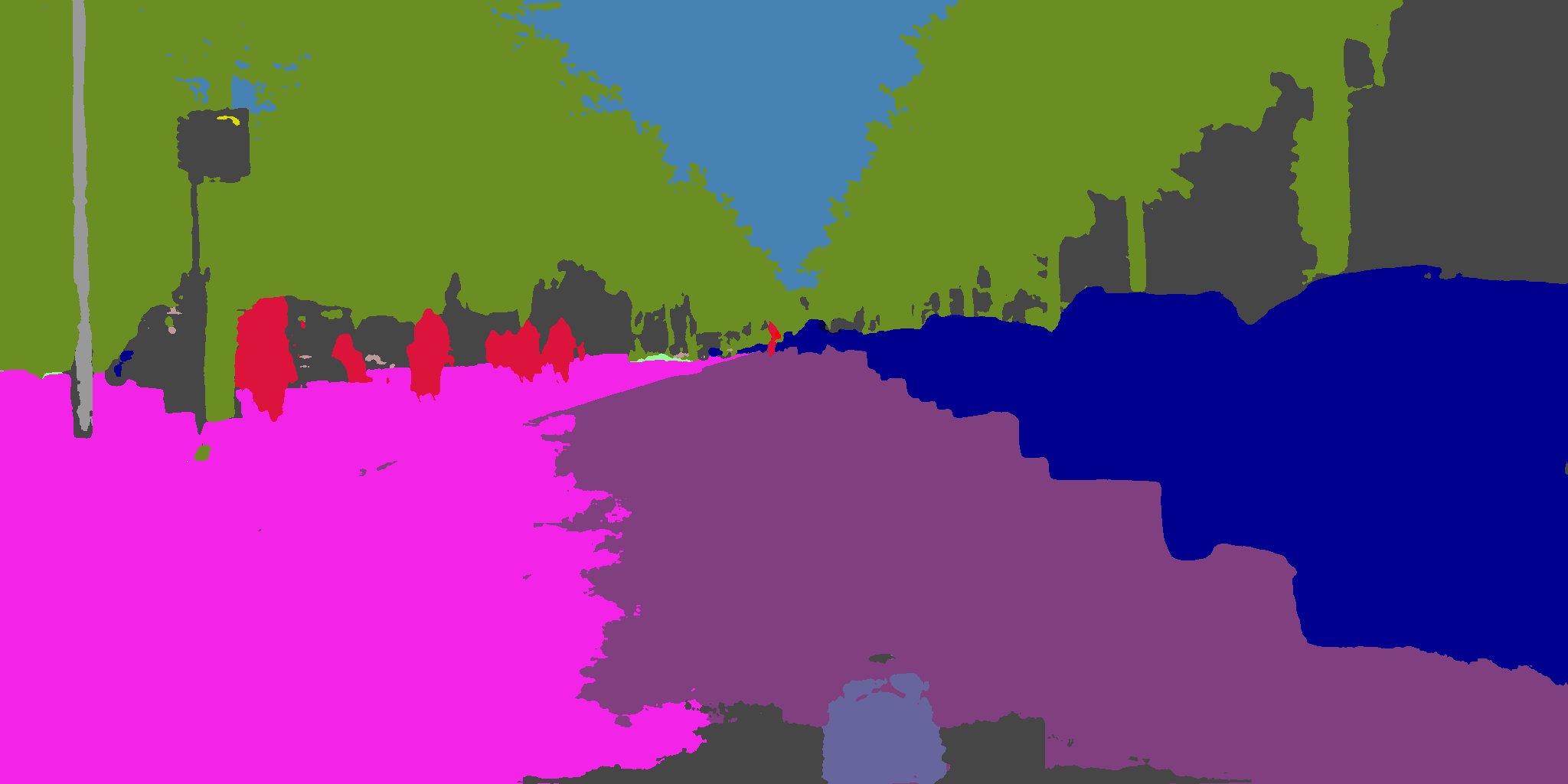}}
    \hfill
    \subfigure[ISW + LCCS]{\includegraphics[width=0.24\linewidth]{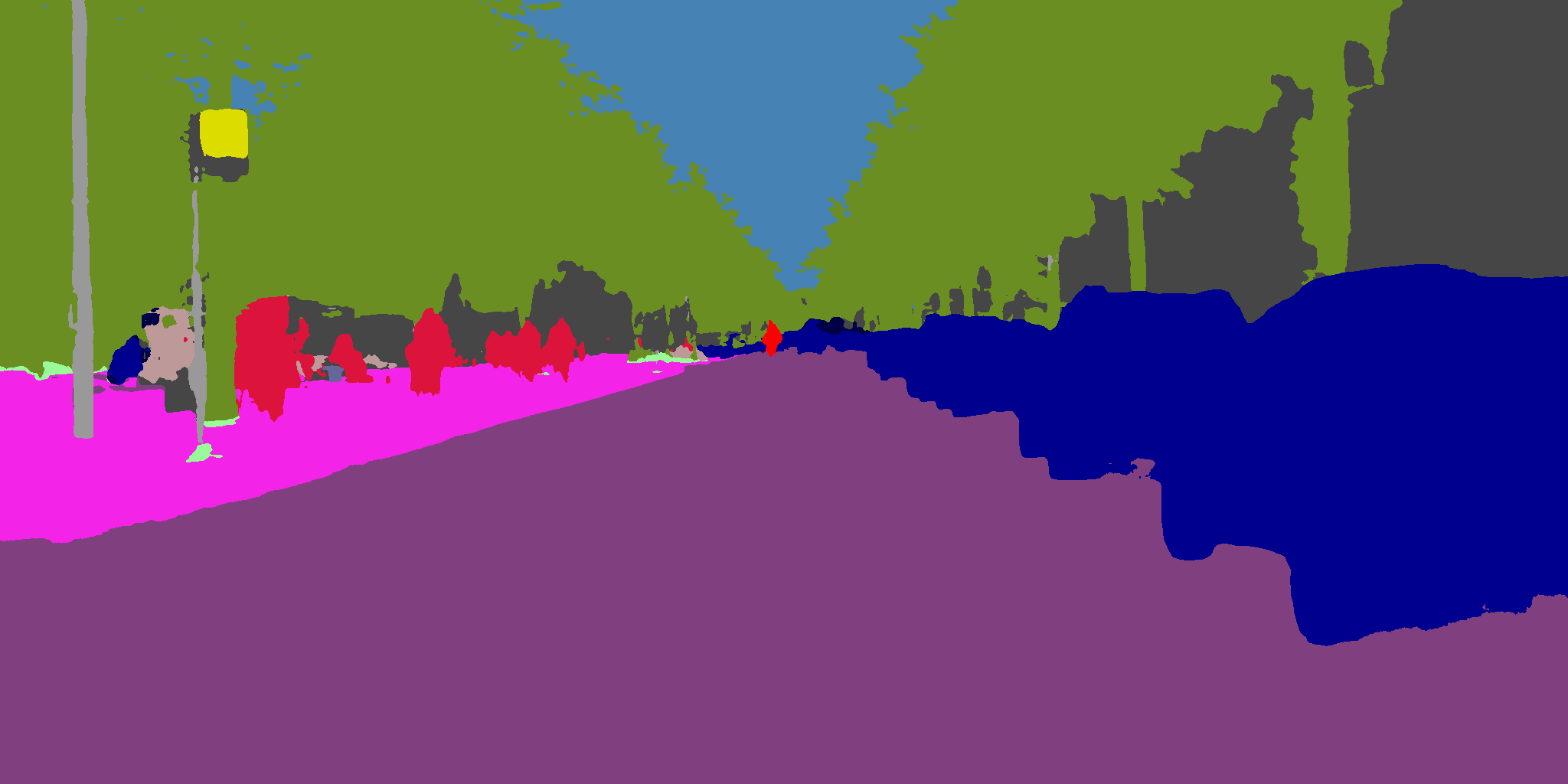}}
    \vspace{-3mm}  
    
    \caption{Semantic segmentation results for GTAV$\rightarrow$ Cityscapes on images from 3 cities: Frankfurt (top), Lindau (middle), and M\"{u}nster (bottom).}
    \label{fig:additional_segmentation_cityscapes}
\end{figure*}

We include additional segmentation visualizations in Figure~\ref{fig:additional_segmentation_cityscapes}. Source dataset is synthetic images from GTAV, target dataset is real-world images from different cities in Cityscapes, and support set is a total of 5 target samples. By adapting ISW pre-trained models~\cite{choi2021isw} with our proposed method, we obtain segmentation visualizations that are more similar to ground truth.

\section{Further Analysis}
\label{sec: further_analysis}

We perform further analysis of the proposed method with the base configuration of $n=1$ and source classifier.

\begin{figure}[hbt]
    \centering
    \includegraphics[width=0.7\columnwidth]{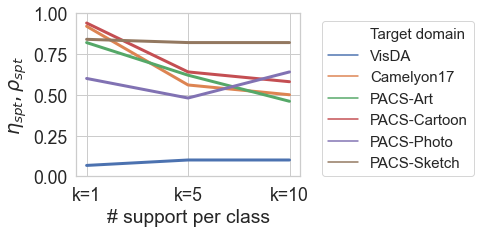}
    \caption{LCCS values selected at initialization stage. \label{fig: lccs_init}} 
    \vspace{-3mm}
\end{figure}

\noindent\textbf{Variable initialization.}
We plot LCCS parameters $\eta_{spt}$ and $\rho_{spt}$ after the initialization stage in Figure~\ref{fig: lccs_init}. Initialization values are shared across all BN layers and chosen by cross-entropy minimization. We observe that initialization tends to be in the middle instead of ends of the $[0,1]$ range, implying that the optimal statistics is not simply source or target statistics. LCCS parameters after gradient update stage deviate slightly from initialization. Deviations differ across BN layers and are typically on the scale of $0.01$ which is small compared to the initialization, hence we do not report these here. iWildCam is not included in Figure~\ref{fig: lccs_init} since only the $k=1$ setting is implemented with limited target data, its chosen initialization for $\eta_{spt}$ and $\rho_{spt}$ is at 0.

\begin{figure}[hbt]
    \centering
    \includegraphics[width=0.5\columnwidth]{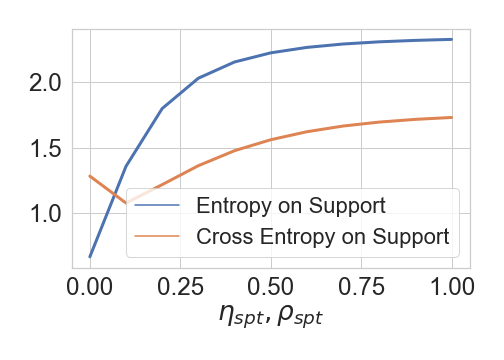}%
    \includegraphics[width=0.5\columnwidth]{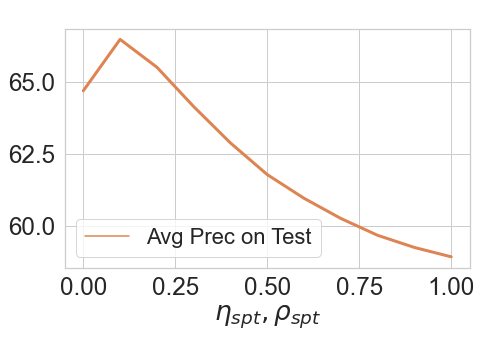}
    \caption{VisDA ($k=1$): Cross entropy, compared to entropy, on support set is more indicative of test performance.}
    \label{fig:visda_diff_init}
    \vspace{-3mm}
\end{figure}

\noindent\textbf{Cross entropy objective.}
Zooming into VisDA with $k=1$ in Figure~\ref{fig:visda_diff_init}, we see that on a grid of candidate values, LCCS with lowest cross-entropy on support set also maximizes test performance. Entropy minimization used in previous works results in lower test performance, hence we advocate that a small amount of labeled target samples are necessary for adaptation performance maximization.

\noindent\textbf{Choice of classifier.}
It is conventional to update the output classifier when sufficient support samples are available. In Table~\ref{tab:classifier_choice}, we provide classification performance with pre-trained source classifier, finetuned classifier (source classifier finetuned for 10 epochs), and nearest-centroid classifier, applied with LCCS parameter finetuning with $n=k\times \text{\# classes}$. We observe that the source classifier is more reliable at $k=1$, and it becomes more beneficial to adapt the classifier with larger support sets. On the 3 datasets evaluated, the nearest-centroid classifier has best performance in most cases at $k\geq 5$.
\begin{table}[htb!]
\centering
\begin{adjustbox}{max width=\linewidth}
\begin{tabular}{*{10}{l}}
\toprule[1pt]\midrule[0.3pt]
\textbf{Classifier}             & \multicolumn{3}{c}{\textbf{PACS}}  & \multicolumn{3}{c}{\textbf{Camelyon17}}  & \multicolumn{3}{c}{\textbf{VisDA}} \\ \cmidrule{2-10}
\multicolumn{1}{r}{$k=$}    & \multicolumn{1}{c}{1} & \multicolumn{1}{c}{5} & \multicolumn{1}{c}{10}
                            & \multicolumn{1}{c}{1} & \multicolumn{1}{c}{5} & \multicolumn{1}{c}{10}
                            & \multicolumn{1}{c}{1} & \multicolumn{1}{c}{5} & \multicolumn{1}{c}{10} \\ \cmidrule(lr){2-4} \cmidrule(lr){5-7} \cmidrule(lr){8-10}
Pre-trained          & \textbf{84.4} & \underline{86.2} & \underline{86.7}            & \underline{76.6} & \textbf{88.6} & \underline{88.9}                & \textbf{67.8} & 69.5 & 72.5 \\
Finetuned          & \underline{83.7} & 86.1 & 86.6            & \textbf{77.1} & \underline{88.4} & 88.6                & \underline{64.3} & \underline{71.1} & \underline{77.7} \\
Nearest-centroid         & 75.2 & \textbf{87.1} & \textbf{88.8}            & 72.4 & 88.3 & \textbf{90.2}                & 52.9 & \textbf{76.0} & \textbf{79.2} \\
\midrule[0.3pt]\bottomrule[1pt]
\end{tabular}
\end{adjustbox}
\caption{Classification performance with different classifiers: fully-connected pre-trained source classifier, fully-connected finetuned classifier on support set, and nearest-centroid classifier on support set. \label{tab:classifier_choice}}
\vspace{-3mm}
\end{table}

\noindent\textbf{Computational Cost.}
Our proposed method has the same inference time as the original network since the introduced LCCS parameters are frozen and absorbed into the BN statistics.

During training, our method imposes minimal extra computation time compared to the original network. For simplicity, consider one epoch of gradient-based training for $N$ samples on a network with one BN layer having $C$ channels. Time complexity for our proposed BN layer is $\mathcal{O}(N(A+Cn))$ for $n$ target domain spanning vectors and constant $A$ denoting time complexity of backpropagating from loss term to BN layer output, versus $\mathcal{O}(N(A+C))$ for the original BN layer. We expect $A >> C$ for large neural networks such that the additional computation cost is relatively small. Empirically on VisDA using a Tesla V100-SXM2 GPU, the average training time per epoch on a vanilla ResNet-101 is 3.18s, 3.56s and 4.32s for $k=1$, 5 and 10 respectively. With our proposed LCCS parameters with default $n=k\times\text{\# classes}$, average times are 3.48s, 4.43s and 6.40s.

In detail, our proposed BN layer is defined by Equations~\ref{eqn: lccs_general_mu} and \ref{eqn: lccs_general_sigma} where $\{\boldsymbol{\eta},\boldsymbol{\rho}\}$ are learnable, and $\mathbf{Z_{BN}} = \mathbf{\check{Z}} \boldsymbol{\gamma} + \boldsymbol{\beta}$ where $\mathbf{\check{Z}} = \frac{\mathbf{Z}-\boldsymbol{\mu_{LCCS}}}{\boldsymbol{\sigma_{LCCS}}}$. 
In the breakdown of time complexity, forward propagation is $\mathcal{O}(C(N+n))$. For backpropagation with loss $L$, note $\frac{\partial L}{\partial \boldsymbol{\eta}} = \sum_{i=1}^N \frac{\partial L}{\partial \mathbf{Z_{BN}^{(i)}}} \frac{\partial \mathbf{Z_{BN}^{(i)}}}{\partial \mathbf{\check{Z}^{(i)}}} \frac{\partial \mathbf{\check{Z}^{(i)}}}{\partial \boldsymbol{\eta}}$. Since $\frac{\partial L}{\mathbf{\partial Z_{BN}^{(i)}}}$ depends on previous layers in the backpropagation, let $\mathcal{O}(A)$ denote its time complexity. By computing the time complexity of each operation, $\frac{\partial L}{\partial \boldsymbol{\eta}}$ (and similarly $\frac{\partial L}{\partial \boldsymbol{\rho}}$) takes $\mathcal{O}(N(A+Cn))$, and gradient update is $\mathcal{O}(n)$.

\end{document}